\documentclass[manuscript]{acmart}

\usepackage{subcaption}
\AtBeginDocument{%
  }

\begin{document}

\title{A Genealogy of Foundation Models in Remote Sensing}

\author{Kevin Lane}
\email{kevin.lane@colorado.edu}
\author{Morteza Karimzadeh}
\email{karimzadeh@colorado.edu}
\affiliation{%
  \institution{University of Colorado, Boulder}
  \city{Boulder}
  \state{Colorado}
  \country{USA}
}

\renewcommand{\shortauthors}{Lane and Karimzadeh}

\begin{abstract}
  Foundation models have garnered increasing attention for representation learning in remote sensing.  Many such foundation models adopt approaches that have demonstrated success in computer vision with minimal domain-specific modification.  However, the development and application of foundation models in this field are still burgeoning, as there are a variety of competing approaches for how to most effectively leverage remotely sensed data.  This paper examines these approaches, along with their roots in the computer vision field.  This is done to characterize potential advantages and pitfalls, while outlining future directions to further improve remote sensing-specific foundation models. We discuss the quality of the learned representations and methods to alleviate the need for massive compute resources.  We first examine single-sensor remote foundation models to introduce concepts and provide context, and then place emphasis on incorporating the multi-sensor aspect of Earth observations into foundation models.  In particular, we explore the extent to which existing approaches leverage multiple sensors in training foundation models in relation to multi-modal foundation models.  Finally, we identify opportunities for further harnessing the vast amounts of unlabeled, seasonal, and multi-sensor remote sensing observations.
\end{abstract}


%%
\begin{CCSXML}
<ccs2012>
   <concept>
       <concept_id>10002944.10011122.10002945</concept_id>
       <concept_desc>General and reference~Surveys and overviews</concept_desc>
       <concept_significance>500</concept_significance>
       </concept>
   <concept>
       <concept_id>10010147.10010178.10010224.10010240.10010241</concept_id>
       <concept_desc>Computing methodologies~Image representations</concept_desc>
       <concept_significance>500</concept_significance>
       </concept>
   <concept>
       <concept_id>10010147.10010178.10010224.10010245.10010254</concept_id>
       <concept_desc>Computing methodologies~Reconstruction</concept_desc>
       <concept_significance>300</concept_significance>
       </concept>
   <concept>
       <concept_id>10010405.10010432.10010437.10010438</concept_id>
       <concept_desc>Applied computing~Environmental sciences</concept_desc>
       <concept_significance>500</concept_significance>
       </concept>
   <concept>
       <concept_id>10002951.10003227.10003236</concept_id>
       <concept_desc>Information systems~Spatial-temporal systems</concept_desc>
       <concept_significance>500</concept_significance>
       </concept>
 </ccs2012>
\end{CCSXML}

\ccsdesc[500]{General and reference~Surveys and overviews}
\ccsdesc[500]{Computing methodologies~Image representations}
\ccsdesc[300]{Computing methodologies~Reconstruction}
\ccsdesc[500]{Applied computing~Environmental sciences}
\ccsdesc[500]{Information systems~Spatial-temporal systems}
\keywords{Foundation Models, Remote Sensing, Self-Supervised Learning}

%TODO: add correct dates here
\received{16 April 2025}
\received[revised]{23 October 2025}
\received[revised]{7 January 2026}
\received[accepted]{12 January 2026}

\maketitle

\section{Introduction}
With the recent successes of Natural Language Processing (NLP) frameworks like Generative Pre-trained Transformers (GPT) \cite{openai_gpt-4_2024} and Bi-directional Encoder Representations from Transformers (BERT) \cite{devlin_bert_2019}, the development and application of machine learning is shifting away from traditional supervised learning.  Supervised learning on a large, labeled dataset learns representations of that data.  These representations are most applicable to the task they have trained on, but can be transferred to other similar tasks to speed up training.  Instead, Self-Supervised Learning (SSL), which learns general-purpose representations of data without the need for external labels, has risen in popularity  \cite{chen_self-supervised_2024}.  As datasets have grown larger, with some datasets now containing billions of data points \cite{dehghani_scaling_2023}, the process of labeling data for supervised learning has proven to be a costly bottleneck for improving model performance.  SSL circumvents this issue with the use of pretext tasks: auxiliary learning objectives designed to leverage the natural structure of unlabeled data in order to learn patterns or relationships within the data.  Researchers can then fine-tune these pre-trained models for specific downstream tasks on significantly smaller labeled datasets.  Fine-tuning enables downstream users to achieve state-of-the-art (SOTA) performance on a specific task while incurring significantly lower costs than training an equivalent model from scratch \cite{anisuzzaman_fine-tuning_2025}.  Thus, effective SSL training democratizes access to high performance machine learning models.

In the wake of NLP success, the computer vision (CV) field swiftly delved into SSL with foundation models such as BYOL \cite{grill_bootstrap_2020}, SimCLR \cite{chen_simple_2020}, MoCo \cite{he_momentum_2020}, and DINO \cite{caron_emerging_2021}, all of which use unique approaches to create self-supervised tasks and objectives for natural imagery data. These methodologies have recently started to percolate into the field of remote sensing (RS), where the sheer amount of unlabeled data from satellite imagery makes the field ripe for the development of foundation models.  Satellite imagery can be used for environmental monitoring tasks, such as monitoring or forecasting wildfires.  Climate change's impact on such events means that there is imminent use for these generalized representations for downstream tasks.  

Thus far, the self-supervised methods for foundation models in RS build heavily upon CV methods and fall broadly into one of four categories:
\begin{enumerate}
    \item Contrastive learning via negative sampling
    \item Contrastive learning via distillation
    \item Contrastive learning via redundancy reduction
    \item Masked image modeling
\end{enumerate}

Most of these approaches have demonstrated significant promise in the RS domain but have largely remained limited to applying the same CV frameworks to satellite imagery and other remote sensing datasets with few to no methodology changes \cite{rolf_mission_2024}.  However, recent research demonstrates how satellite data has unique properties that are distinct from natural imagery, such as the logarithmic distribution of objects and the range of the electromagnetic spectrum captured within satellite images \cite{rolf_mission_2024}.  These properties indicate that satellite imagery is a distinct modality and would therefore benefit from modality-specific frameworks and self-supervised learning tasks.  Moreover, the variety of different satellite and ground-level sensors available allows for the development of highly specialized multi-modal foundation models.  In this paper, we contribute to the growing body of research on foundation models in remote sensing by doing the following:
\begin{enumerate}
    \item We further demonstrate how the modes of data available in RS provide unique information that sets them apart from the natural imagery of CV.
    \item We survey recent foundation models developed in the field, categorized by their learning objective, and discuss their specific adaptations for RS observations.
    \item We identify gaps and outline research opportunities for RS foundation models.  We specifically consider the multi-sensor and multi-modal nature of RS observations, and how to effectively leverage such data at scale.
\end{enumerate}

\section{Prior Work}
Recent papers in the field have surveyed available foundation models in RS.  Zhou et al. explored the high potential of applied Vision Language Models (VLMS) while highlighting their few shot capabilities in downstream tasks \cite{zhou_towards_2024}.  Chen et al. inspected how models can adapt to utilize non-raster based data \cite{chen_self-supervised_2024}, such as poly-lines, to represent transportation networks.  Zhang et al. examined the singular task of building foundation models specifically for urban environments \cite{zhang_urban_2024}.  Mai et al. created a 'wish list' of desirable traits in RS foundation models, with rationale for each wish, and discussed where current foundation models exceed or fall short of these goals \cite{mai_opportunities_2023}. Lu et al. categorized the model backbones, training datasets, image resolution, and over-arching pre-training methods for existing RS foundation models \cite{lu_vision_2025}.  Lu et al. also documented the number of parameters for each model and compiled results across a wide range of diverse downstream tasks \cite{lu_vision_2025}.  

While these works provide a timely review of important components in RS foundation models, none of them focus on the specific adaptations that have been made to SSL approaches for working with remote sensing data.  Works such as \cite{zhu_foundations_2024} might note that ideal RS foundation models should factor in temporal features, but do not spend time exploring the ins-and-outs of how current models do so.  Surveys such as \cite{lu_vision_2025} comprehensively document the different datasets, backbones, and broad SSL tasks used by RS foundation models.  However, they do not delineate the differences in models past this point, preferring to focus instead on model performance.  While comprehensive benchmarking is helpful, differences in models' learning objectives such as the use of temporally positive pairs, adaptation to multi-scale view, or leveraging a remote sensing index as a reconstruction target, are all glossed over.  These kinds of adaptations for RS data inform model performance and effective downstream use, and more importantly, future research directions.  We  fill that gap here by discussing these unique adaptations.

\begin{figure}[ht]
    \centering
    % First Row of Images
    \begin{subfigure}{0.3\textwidth}
        \centering
        \includegraphics[width=\linewidth]{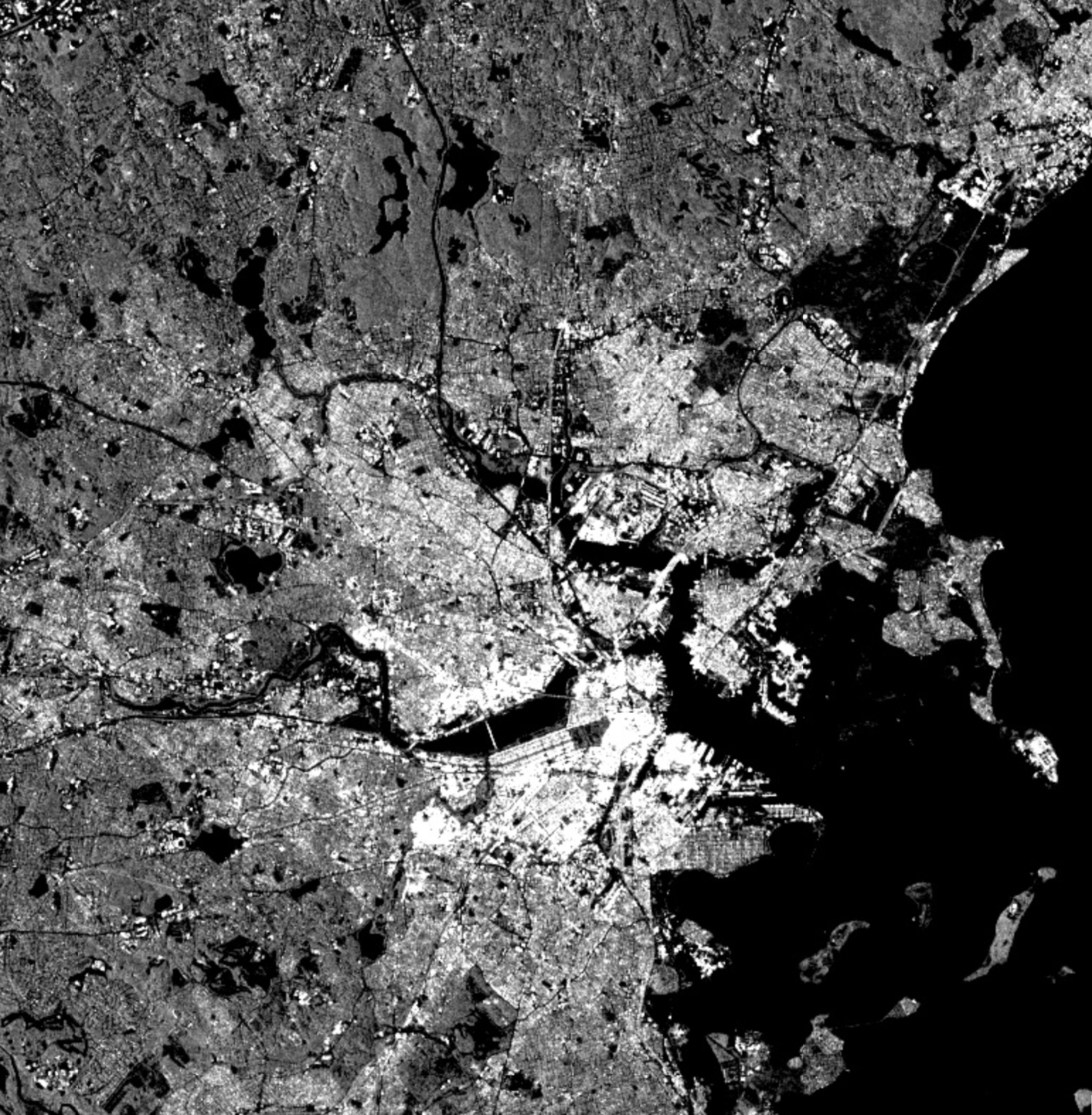}
        \caption{SAR VH polarization}
        \label{fig:vh_polarization}
    \end{subfigure}%
    \hspace{0.03\textwidth} % Space between the images
    \begin{subfigure}{0.3\textwidth}
        \centering
        \includegraphics[width=\linewidth]{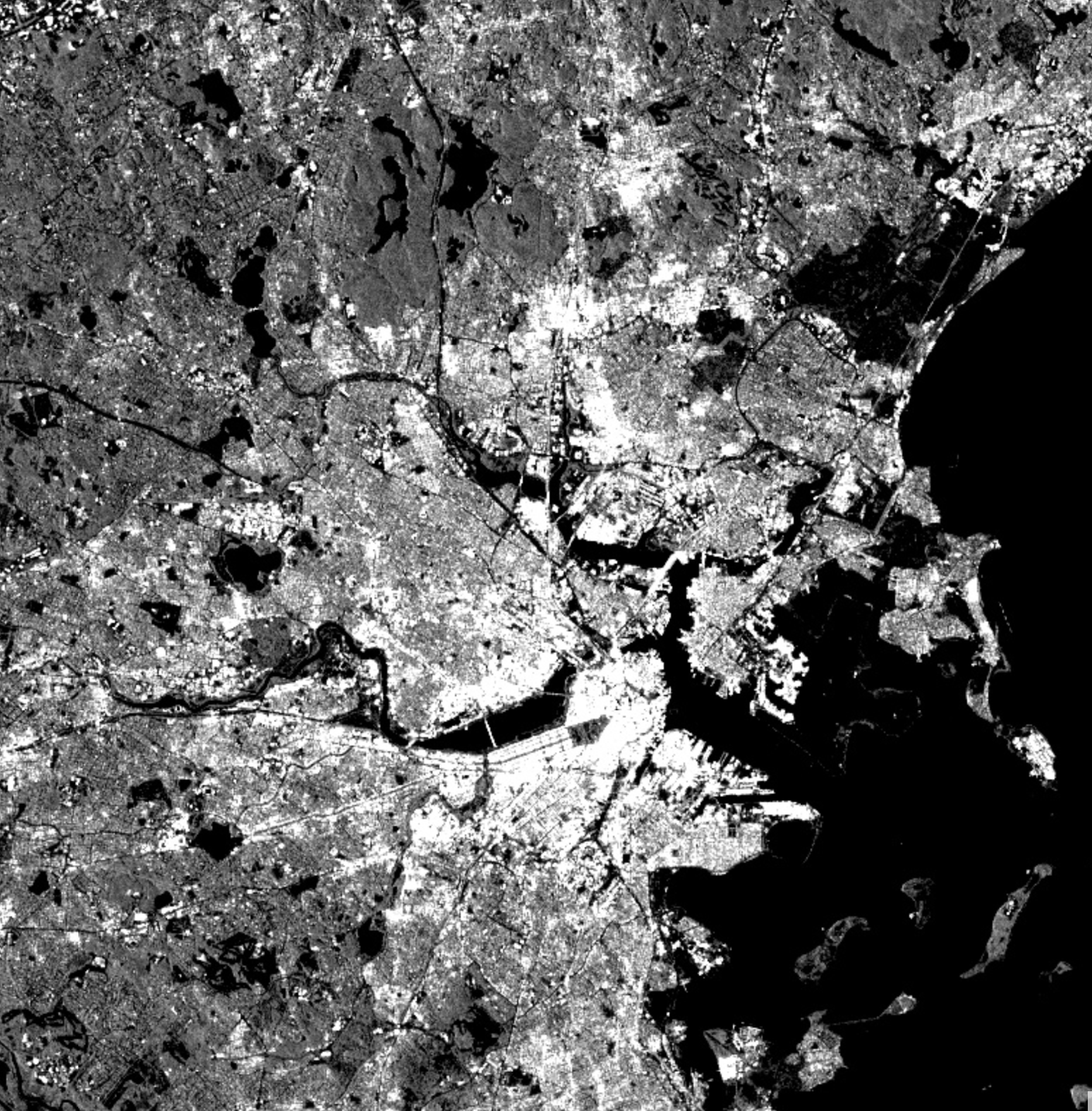}
        \caption{SAR VV polarization}
        \label{fig:vv_polarization}
    \end{subfigure}%
    \hspace{0.03\textwidth} % Space between the images
    \begin{subfigure}{0.3\textwidth}
        \centering
        \includegraphics[width=\linewidth]{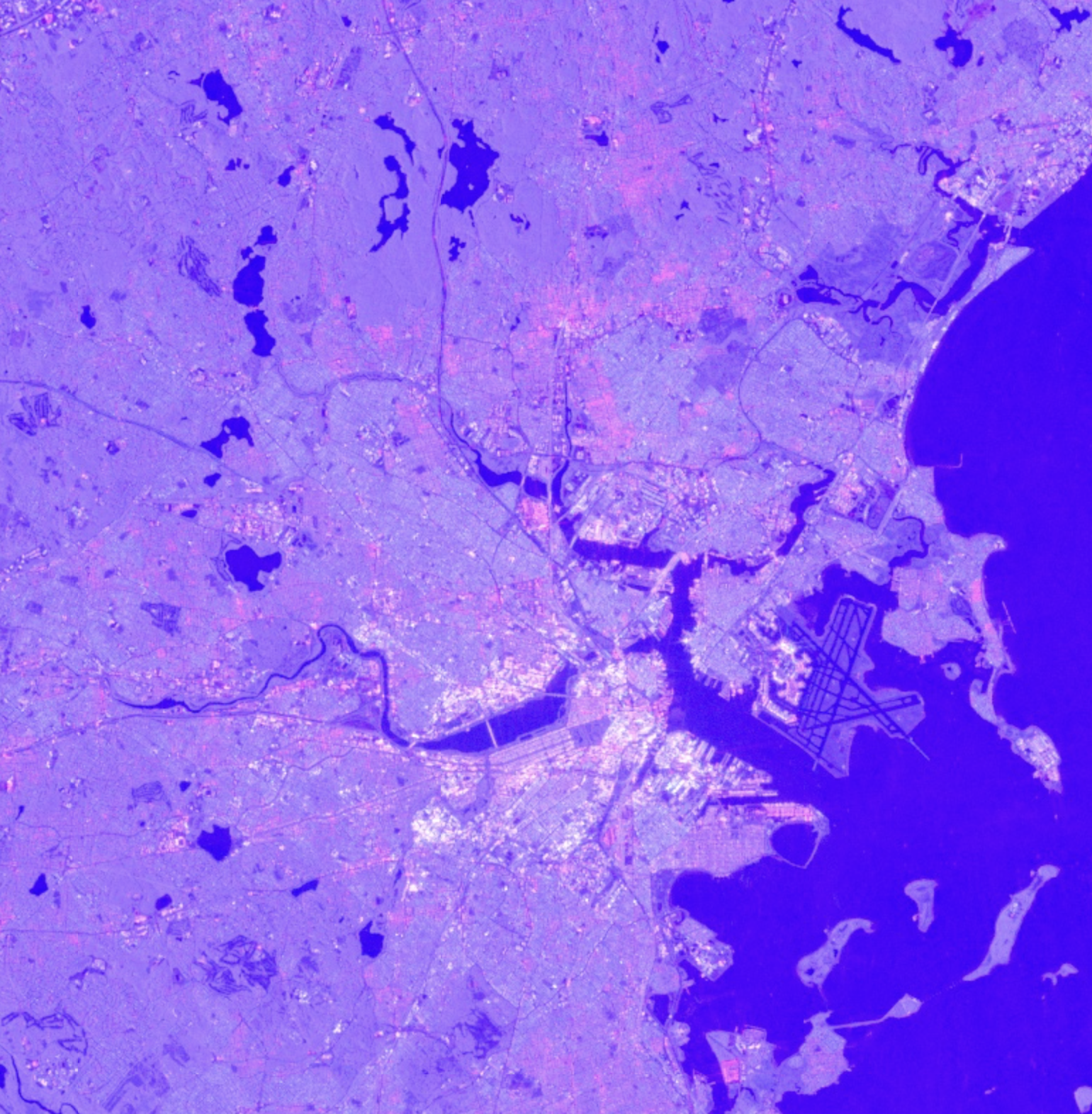}
        \caption{SAR False Color}
        \label{fig:sar_synthetic}
    \end{subfigure}

    \vspace{0.05\textwidth} % Vertical space between the rows

    % Second Row of Images
    \begin{subfigure}{0.3\textwidth}
        \centering
        \includegraphics[width=\linewidth]{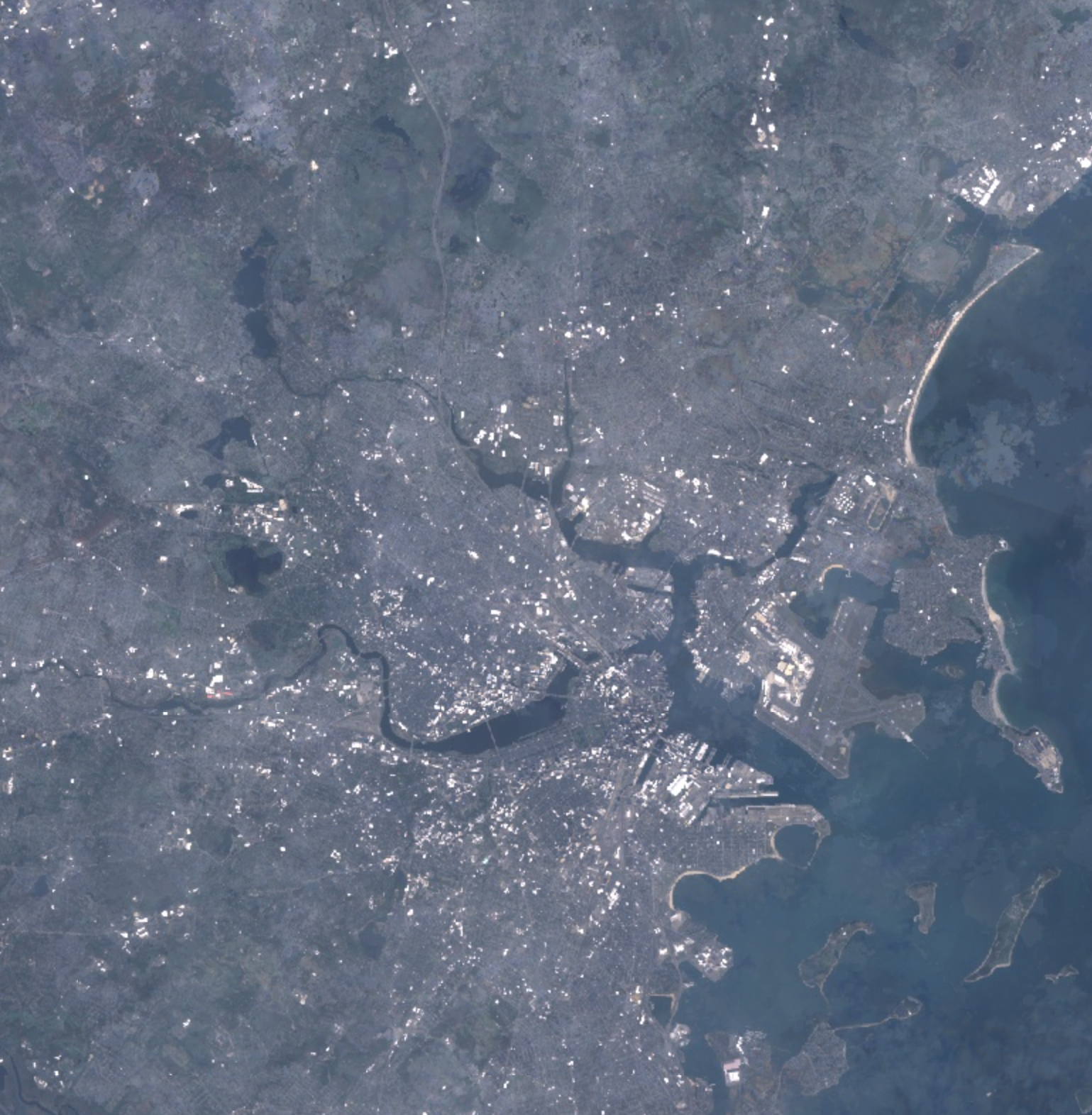}
        \caption{Optical RGB}
        \label{fig:optical_rgb}
    \end{subfigure}%
    \hspace{0.03\textwidth} % Space between the images
    \begin{subfigure}{0.3\textwidth}
        \centering
        \includegraphics[width=\linewidth]{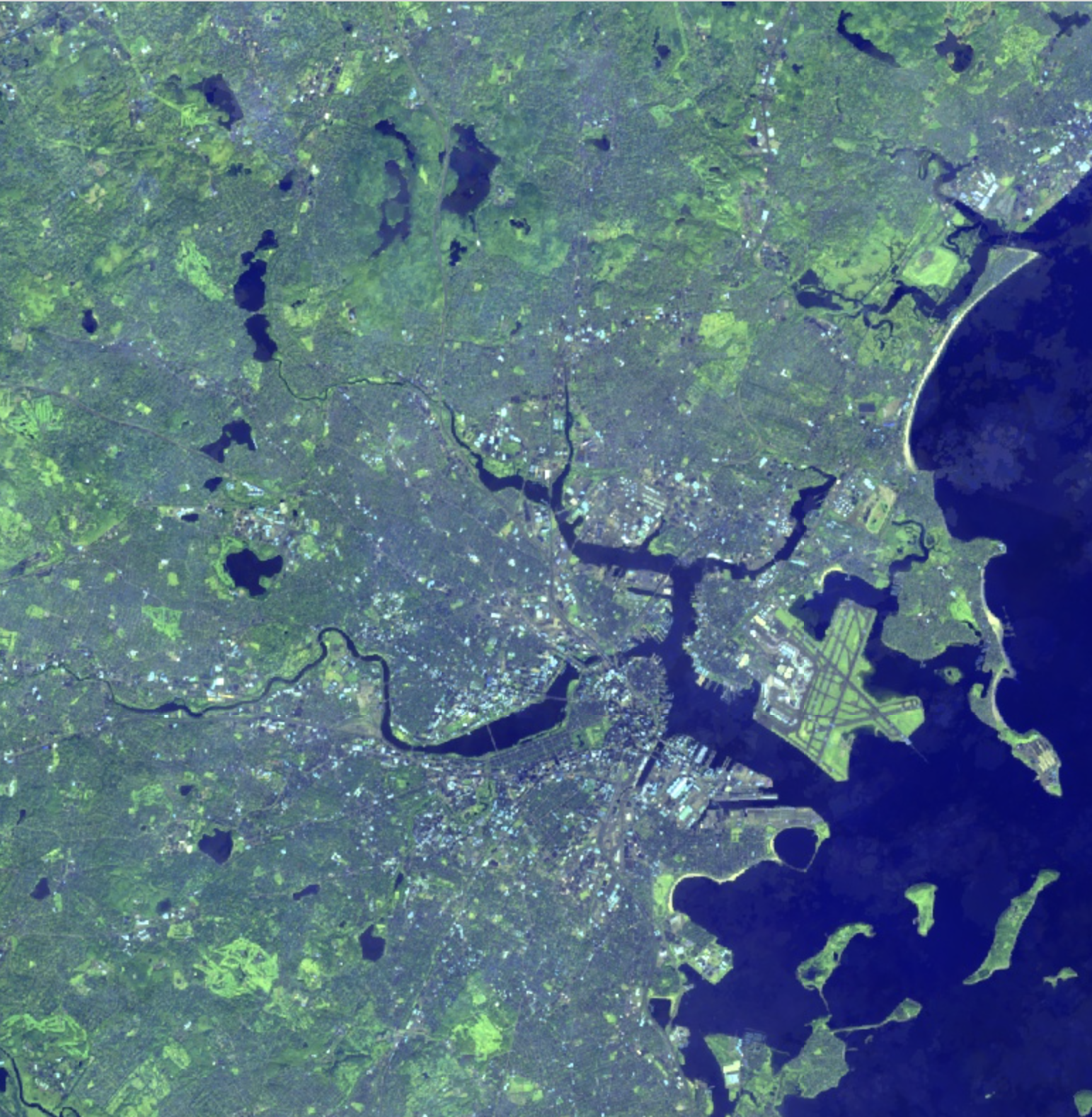}
        \caption{Optical SWIR \& Blue}
        \label{fig:optical_geology}
    \end{subfigure}%
    \hspace{0.03\textwidth} % Space between the images
    \begin{subfigure}{0.3\textwidth}
        \centering
        \includegraphics[width=\linewidth]{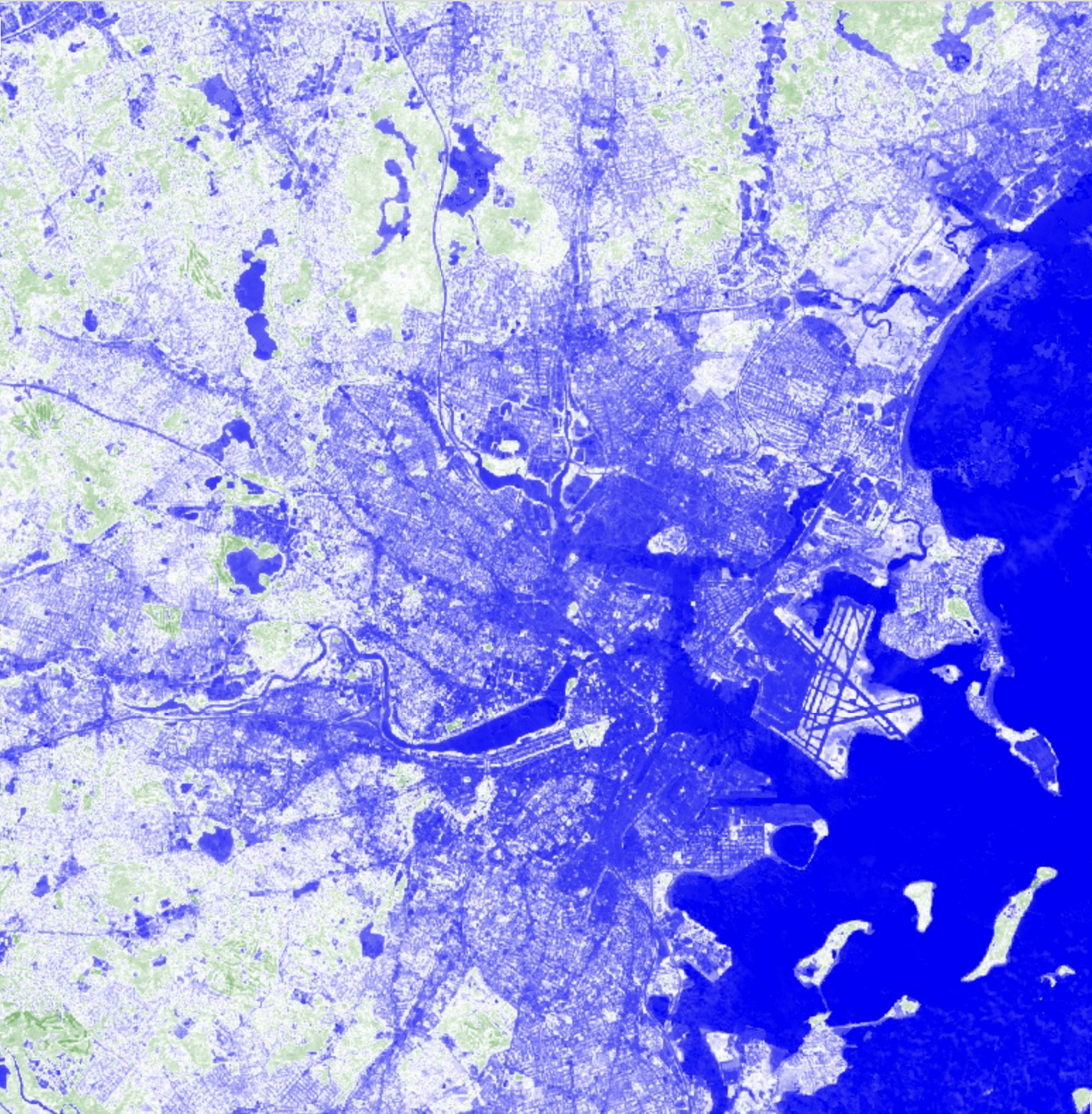}
        \caption{Optical NDVI}
        \label{fig:optical_ndvi}
    \end{subfigure}

    \caption{Satellite imagery of Boston, MA in October, 2020. SAR imagery is obtained from Sentinel-1, and optical imagery is obtained from Sentinel-2.  SAR false color uses VV polarization for red channel, VH polarization for green channel, and the average of VV and VH polarizations for blue channel.  NDVI refers to the Normalized Difference Vegetation Index, an index based on the Red and Near Infrared (NIR) bands that is utilized to assess vegetation health.}
    \Description{A variety of satellite images capturing Boston, MA in the month of October, 2020.  Images with VV and VH polarizations captured by Synthetic Aperture Radar (SAR) are in black and white and do a good job of capturing water and urban structures.  A false color image of SAR combines the different polarizations to provide a sharper, 3-channel image.  There is a standard RGB optical image, and a false color image with Shortwave Infrared (SWIR) bands and the blue light band that is used to highlight geological features.  Then, there is a depiction of NDVI, where dark green highlights healthy vegetation, dark blue highlights no vegetation, and white indicates sparse or unhealthy vegetation.}
    \label{fig:satellite images}
\end{figure}

\section{Data for Training  Remote Sensing Foundation Models}
The CV field has primarily been concerned with natural images, which only sense the red, green, and blue wavelengths of the visible spectrum, and are typically captured at ground level.  Some models, such as CLIP \cite{radford_learning_2021}, FLIP \cite{jia_scaling_2021}, VisualBERT \cite{li_visualbert_2019}, ViLBERT \cite{lu_vilbert_2019}, FLAVA \cite{singh_flava_2022}, and DeepSeek-VL2 \cite{wu_deepseek-vl2_2024} are VLMs that also utilize text as an additional mode of data.  However, most of the RS foundation model frameworks borrowed from CV concern themselves with only the image modality.  Within the RS domain, the variety of unique sensors available ranges across a number of instruments aboard different satellites.  The data modes detailed below are available at mass scale and readily accessible for training foundation models.  The training of RS foundation models is our focus in this manuscript, rather than providing a full catalogue of remote sensing or geospatial data. We will keep discussions focused on sensors with high resolution (e.g., less than 40m), which are also the focus of current efforts in benchmarking datasets \cite{lacoste_geo-bench_2023} and training foundation models \cite{zhou_towards_2024}.

\subsection{Optical Imagery}
Optical images capture electromagnetic wavelengths in the visible and near visible bands of light.  These are recorded via passive sensors, which simply measure the electromagnetic radiation originating from the sun that is reflected off of objects.  This means that passive sensors are more energy efficient than active sensors, but they are affected by light and weather conditions.  These sensors can only capture imagery during the daytime, and can be obscured by atmospheric phenomena like cloud cover or haze.  This means their use is limited in hurricanes, storms or active floods under overcast conditions.

Optical images are often acquired from different satellites.  Several datasets from different satellites are openly available for free, mass download, such as MODIS \cite{justice_overview_2002}, the Landsat \cite{wulder_fifty_2022} family, and Sentinel-2 \cite{spoto_overview_2012}, with the latter two offering high resolution.  Each satellite contains a sensor with different spatial and spectral resolutions, but each is broadly designed to capture the visible, near-infrared, and short-wave infrared portions of the electromagnetic spectrum.  In the case of the Sentinel-2 satellite, there are 13 distinct bands that are captured at resolutions ranging from 10m/pixel to 60m/pixel.  These bands can be broken down into five groups based on usage as follows \cite{wang_fusion_2016}:
\begin{itemize}
    \item B1 is the coastal aerosol band, useful for mapping coastal marine habitats and bathymetry. \cite{poursanidis_use_2019}
    \item B2-4 capture  RGB (Red Green Blue), similar to digital cameras. 
    \item B5-8 refer to VNIR (Visible and Near Infrared) light, and are typically used to measure vegetation health. B8a is a Narrow Near Infrared (NIR) band for moisture-based vegetation health monitoring. 
    \item B9, still in NIR spectrum, focuses on water vapor absorption and is used to measure atmospheric moisture presence.  It is paired with B8A to derive total precipitable water.
    \item B10-B11 are SWIR (Short Wave Infrared) bands.  B10 is used for cirrus cloud detection. B11 is used for soil and vegetation moisture detection and snow/ice vs. cloud differentiation.  B12 is often used with B2 to measure geological features and map burned areas.
\end{itemize}

This range of spectrum and information captured by Sentinel-2 is an example of the wealth of information captured by remote sensing optical products that go beyond RGB. However, CV models are designed to operate with three input channels of RGB of ‘natural’ images (obtained by regular cameras and widely available on the internet).  In order to leverage the highly sophisticated pre-trained weights that are available for CV backbones, such as the ubiquitous ResNet50 \cite{he_deep_2015} or ViT16 \cite{dosovitskiy_image_2021}, some RS models only utilize the visible bands of light from optical multispectral (MS) satellite images \cite{manas_seasonal_2021, chen_mmcpp_2023}.  Some RS models also rely on RGB images taken at ground level \cite{cepeda_geoclip_2023}, or from some sort of aerial platform, such as a drone or helicopter \cite{liu_remoteclip_2024}.  While RS models trained purely on RGB can achieve SOTA performance in certain settings \cite{manas_seasonal_2021}, deep learning models in RS that only train on RGB data have an information disadvantage compared to models that train on all bands.  RGB-only models can struggle to perform on a number of downstream tasks in RS as a result \cite{lacoste_geo-bench_2023}. The spectral range of interest and the specific choice of bands on a satellite are the result of decades of experience, in-lab experiments and consensus-building efforts to ensure satellites are equipped with bands that can assist in target-specific tasks. Figure \ref{fig:satellite images} highlights the information lost when solely examining RGB imagery. For instance, infrared bands are crucial for monitoring vegetation \cite{neinavaz_thermal_2021}, as seen in figure \ref{fig:optical_ndvi}, while NIR bands in confluence with blue and green bands can help with water quality analysis \cite{cillero_castro_uav_2020}. SWIR can be used for measuring soil and vegetation moisture, and is depicted in figure \ref{fig:optical_geology}.  

Many satellite missions employ mission-specific data pre-processing pipelines and provide multiple standardized data products. For example, Sentinel-2 offers two primary optical products, and different teams have chosen either to train foundation models:
\begin{itemize}
    \item Level-1C (L1C) provides Top-of-Atmosphere (TOA) reflectance for all 13 spectral bands and includes standard radiometric calibration, geometric correction, and ortho-rectification, but does not apply atmospheric correction.\cite{spoto_overview_2012}
    \item Level-2A (L2A) provides Bottom-of-Atmosphere (BOA), or surface reflectance, imagery derived via atmospheric correction. This product excludes B10, which is not suitable for surface reflectance retrieval.\cite{main-knorn_sen2cor_2017}
\end{itemize}

The majority of remote-sensing foundation models discussed in this manuscript that leverage Sentinel-2 data are trained using the L1C product. Unless otherwise specified, models trained on Sentinel-2 data are assumed to use L1C imagery.

While an increasing number of foundation models leverage most bands present in optical imagery such as Sentinel-2, they typically do so by a simple manipulation of input channel dimensions to the image encoder of the model \cite{wang_self-supervised_2022}. The optimal architecture required for remote sensing architectures is not well-defined yet \cite{rolf_mission_2024}.

\subsection{Synthetic Aperture Radar}
Synthetic Aperture Radar (SAR) is an active remote sensor that emits energy in microwave frequencies towards the earth and then captures the amount of energy reflected back.  Capturing the image from a moving sensor (in this case, a satellite) allows for capturing the image through a large 'synthetic' aperture, compared to the much smaller physical sensor aperture.  This results in a much higher resolution than other sensors that monitor similar frequencies \cite{brown_introduction_1969}. Each SAR sensor operates in a specific wavelength, referred to as bands, which serve different uses \cite{tsokas_sar_2022}.  Larger wavelengths provide more penetration through obstruction, such as vegetation, but result in lower resolution.  One of the most commonly used bands is C-band (7.5-3.8 cm wavelength) \cite{tsokas_sar_2022}, which is currently captured by Sentinel-1 \cite{torres_gmes_2012}, RADARSAT-1 \cite{mahmood_radarsat-1_2014}, RADARSAT-2 \cite{morena_introduction_2004}, RADARSAT Constellation Mission (RCM) \cite{thompson_overview_2015} satellites.  C-band imagery is made available publicly, and is useful for global mapping and change detection.  The larger wavelength provides moderate penetration still at relatively high spatial resolution (e.g., 40m).  However, there are larger wavelengths available, such as L-band, which is captured by  AirSAR \cite{hoekman_interpretation_1993}, PALSAR \cite{rosenqvist_alos_2007}, and the soon-to-be-launched NISAR \cite{chapman_overview_2024}.  L-band is often used for biomass and vegetation mapping \cite{haldar_assessment_2012}.  There is also X-band, whose wavelength is smaller than C-band and often used for high resolution monitoring of fine surface features, such as vegetation structure, and urban infrastructure. \cite{solari_combined_2017}.  X-band is obtained by platforms such as Capella \cite{castelletti_capella_2021}, ICEYE \cite{ignatenko_iceye_2020}, TerraSAR-X \cite{werninghaus_terrasar-x_2010}, and TanDEM-X \cite{krieger_tandem-x_2007}. Different wavelengths of SAR all capture different information, and different wavelengths often complement each other if available for the same region and time.

SAR has another characteristic distinct from optical imagery: SAR takes advantage of materials’ innate dielectric properties. Every material has a dielectric constant that describes how strongly it interacts with electromagnetic waves.  Materials with a high dielectric constant, such as water, reflect microwave energy back more strongly, allowing it to be picked up more easily by SAR \cite{barowski_spatial_2019}.  This can prove useful for applications such as mapping waterways in a region that might be otherwise undetected by optical sensors.  As seen in figure \ref{fig:satellite images}, SAR imagery clearly highlights water features that optical sensors struggle to pick up.  SAR is an all-sky sensor, meaning that it operates regardless of light conditions.  Its wavelength range is also large enough that it penetrates cloud cover.  This combination of traits helps provide continuous spatial coverage in geographic areas that might otherwise be occluded to optical satellite sensors due to cloud cover \cite{brown_introduction_1969}, or polar regions that lack sunlight for half of a year. 

SAR is predated by passive microwave sensing, which monitors similar microwave frequencies in order to pierce cloud cover but does not emit a signal of its own in order to do so.  As a result, passive microwave sensing offers much lower resolution compared to SAR.  Sensors such as AMSR \cite{tachi_advanced_1989}, AMSR-2 \cite{alsweiss_inter-calibration_2015}, SMAP \cite{entekhabi_soil_2010}, and GPM \cite{munchak_activepassive_2020} all passively monitor microwave signals emitted from the Earth.  By comparison, SAR actively beams energy down to Earth’s surface and records the response, which allows for multiple polarizations at transmission and reception.  Polarization describes the orientation of the plane that the electromagnetic wave oscillates on.  The shorthand for polarization is the orientation of the emission, which is either vertical (V) or horizontal (H), followed by the orientation of the reception, which will also be either vertical or horizontal.  Each polarization results in the waves scattering differently upon reflection, enabling certain polarizations to be particularly effective at picking up certain types of surface features.  For example, the VV polarization is good at picking up rough surface scattering, typically caused by roughened water or vertical surfaces.  HV or VH polarizations are good at picking up volume scattering, where the signal bounces around a number of times in a given volume before being broadcast back out to space.  This is typically caused by forest canopies and vegetation.  We see in figure \ref{fig:vh_polarization} and figure \ref{fig:vv_polarization} that the VH polarization provides more fine-grained details in a variety of forested areas than the VV polarization, where those areas are whited out. HH polarization is good at picking up horizontal surfaces, such as smooth ice or calm water.

Compared to optical imagery, SAR captures distinct and complementary information, despite it also being stored as a raster (gridded) data type.  The different frequencies of SAR, the different polarizations, and the volume scattering aspect make it stand apart from optical. For instance, surface reflectance of sea ice will appear as a bright surface in optical imagery, regardless of its age (which is a proxy for its thickness and the level of hazard it poses to marine navigation). However, as sea ice ages (from thinner first-year ice to thicker multi-year ice), its salinity changes.  Because of SAR's volume scattering (caused by salt content and air bubbles in ice), the appearance of thick ice changes in SAR imagery compared to younger, thinner ice \cite{pires_de_lima_enhancing_2023}. This is the primary reason sea ice analysts rely on SAR imagery in making sea ice charts and use optical only as a secondary reference. As we demonstrate later in this manuscript, this complementary aspect of SAR imagery (despite it being stored as an image) motivates incorporating this multi-sensor information into RS foundation models. 

\subsection{Non-Image Data}
Due to the ubiquity of the raster image format in remotely sensed data, current RS foundation models tend to be focused on using images to train, but a variety of other modalities do exist.  Text, geographic location, and digital elevation maps (DEM) are all among potential additional modalities available in remote sensing at scale.  It is conceivable that derivative information products such as population movement trajectories, social media data, and vector maps may be incorporated in enhancing foundation models.  Any dataset with sufficient spatial coverage that contains geographic coordinates and a timestamp can feasibly be compared with corresponding satellite imagery.  Such comparisons would further improve a model’s ability to interpret how satellite imagery pertains to ground level information.   This spatiotemporal relationship results in a broad swathe of different modalities available for RS models to leverage.  With an eye on these additional potential modalities, the rest of this manuscript focuses on the present and immediate future of foundation models in remote sensing.

\section{Foundation Models Originating in Computer Vision}
This section presents a broad overview of the techniques used by current foundation models in CV, many of which serve as the bedrock for RS foundation models today \cite{lu_vision_2025}. 

\subsection{Negative Sampling Models}
One successful approach to SSL in CV is contrastive learning via negative sampling \cite{chen_simple_2020}.  Contrastive learning augments one image in two different ways, then a model is trained to recognize whether two images are augmentations of each other, i.e., are a positive pair.  Any images that are not augmentations of one another are defined as negative pairs.  For every batch, an image will have only one positive pair, and n-1 negative pairs, where n is the batch size.   This approach can be thought of as a classification task where the model trains on classifying two augmentations as either a positive or negative pair.  Within a given batch of samples, normalized-temperature cross entropy loss (InfoNCE Loss) \cite{chen_simple_2020} pulls together embeddings from the same original image and pushes apart embeddings that came from different images.  This results in an encoder that learns sophisticated embeddings of images without the need for supervised learning.  The encoder then can be ported to other downstream tasks, while reducing the need for labeled samples. 

SimCLR \cite{chen_simple_2020} achieved great success with this approach.  SimCLR trained on ImageNet ILSVRC-2012 \cite{russakovsky_imagenet_2015} and introduced a non-linear projection head to improve the embedding quality of the layer before it.  They also discovered that augmentations like color jitter and random cropping were particularly effective in forcing the model to learn embeddings that represented the data as a whole.  Training without augmentations resulted in a model that learned dominant features, such as color distribution, and were able to trivialize the SSL task.  Momentum Contrast (MoCo) \cite{he_momentum_2020} also utilizes contrastive learning, but maintains a dynamic memory bank or queue of negative samples, and updates it using a momentum encoder. This approach allows MoCo to effectively utilize smaller batch sizes than SimCLR, reducing memory requirements. Contrastive Language Image Pre-training (CLIP) \cite{radford_learning_2021} is a contrastive learning framework that utilizes image captions as a way to supervise learning image representations. CLIP trains two encoders, one for the text modality and one for the image modality.  Given the multi-modal nature of this approach, contrastive learning to bring positive pairs together results in an alignment of the two modalities in the same embedding space. All of these models introduce concepts that have been leveraged by recent foundation models within the RS domain \cite{wang_self-supervised_2022}.

\begin{figure}[ht]
    \centering
    \begin{subfigure}{0.45\textwidth}
        \centering
        \includegraphics[width=\linewidth]{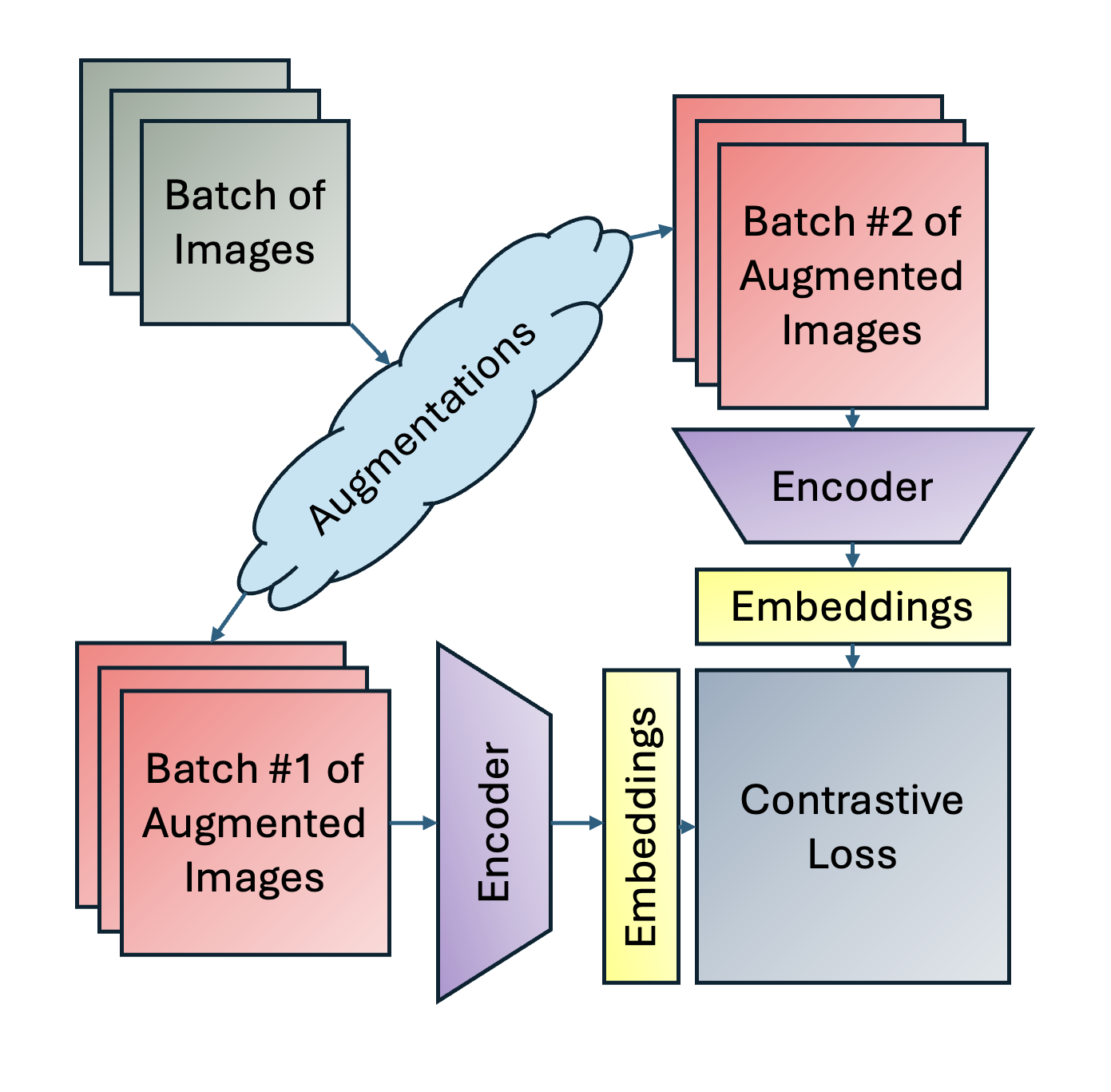}
        \caption{Negative sampling}
        \label{fig:negative_sampling}
    \end{subfigure}%
    \hspace{0.05\textwidth} % Space between the images
    \begin{subfigure}{0.45\textwidth}
        \centering
        \includegraphics[width=\linewidth]{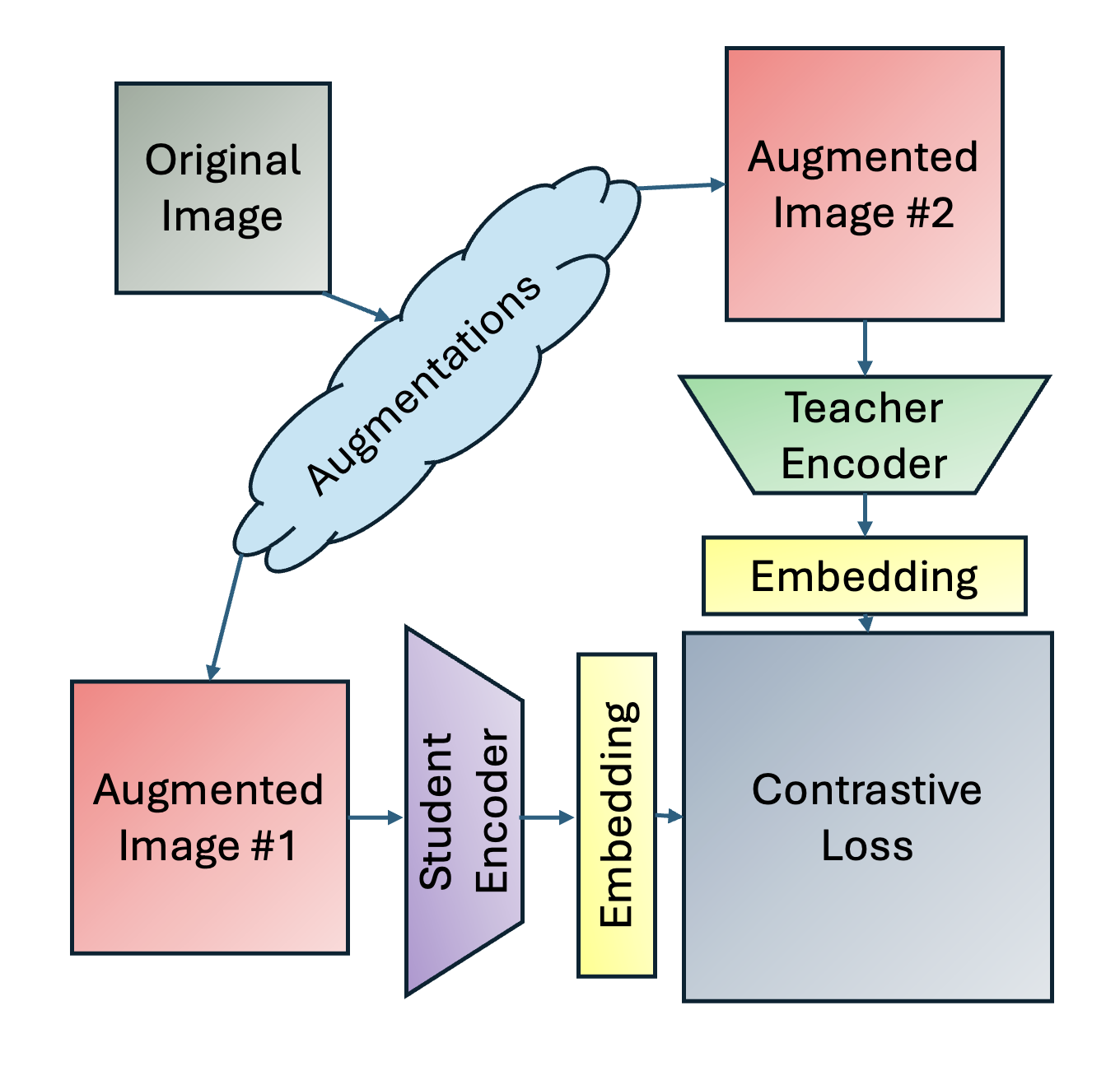}
        \caption{Distillation}
        \label{fig:distillation}
    \end{subfigure}%

    \caption{CV foundation model architectures that leverage contrastive learning.}
    \Description{Two small diagrams that depict contrastive learning with negative sampling or distillation.  In negative sampling, we see a batch of images augmented, passed through the same encoder, and then compared.  In distillation, we see a single image augmented, passed through two different encoders, and then compared.}
    \label{fig:contrastive_diagram}
\end{figure}

\subsection{Distillation Frameworks}
While contrastive learning via negative sampling achieves SOTA performance in CV, it generally requires large batch sizes.  The number of negative pairs increases exponentially with relation to the batch size, and negative pairs are a key component of InfoNCE loss.  This need for a large batch size results in significant hardware requirements, which is an obstacle to research for many teams.  Distillation networks such as Bootstrap Your Own Latent (BYOL) \cite{grill_bootstrap_2020} and Self-Distillation with No Labels (DINO) \cite{caron_emerging_2021} were developed in part to address those hardware constraints.

Distillation simply refers to the concept of one model being trained on the outputs of another model \cite{phuong_towards_2019}.  Distillation is not  strictly an SSL method, as some research focuses on using distillation to 'compress' knowledge from a large pre-trained model into a smaller model \cite{wu_deepseek-vl2_2024}.  Distillation leverages two encoders, dubbed the “teacher” and the “student”.  When trying to 'compress' logic, the teacher is some pre-trained model, but in the context of SSL, both student and teacher encoders are learning, albeit not at the same rate or with the same method, depending on the specific set up.  Both are fed in augmented versions of an image to output embeddings.  Then, contrastive loss is performed between the teacher and student embeddings and back propagated through the student network.  The teacher network typically learns via the weighted average update system that was introduced in MoCo \cite{he_momentum_2020}.  The above broadly describes the framework of BYOL \cite{grill_bootstrap_2020}.  DINO \cite{caron_emerging_2021} has a very similar framework, but utilizes basic centering and sharpening operations on the output from the teacher to avoid embedding collapse.  Distillation models such as Joint-Embeddings Predictive Architecture (JEPA) \cite{assran_self-supervised_2023} build on this framework by introducing a decoder after the student to predict the embeddings from the teacher.  This approach leverages reconstruction loss in the latent space, but more distillation SSL approaches rely on contrasting embeddings from the student and teacher encoders \cite{grill_bootstrap_2020, chen_comprehensive_2022, jang_self-distilled_2023, oquab_dinov2_2024}.

The distillation methodology reduces the reliance on large batch sizes and extensive memory, making it more resource-friendly.  It also outperforms comparable negative sampling models with fewer parameters \cite{grill_bootstrap_2020}.   However, there is possibility of embedding collapse, as both the teacher and student networks learn based on the same loss \cite{chen_exploring_2021}.  Embedding collapse describes a condition where a model generates embeddings that span a low-dimension subspace of the latent space, sometimes becoming so extreme as to only span a single point; this represents a significant loss of ability to capture information in embeddings \cite{jing_understanding_2022}.   Frameworks introduce asymmetry through weighted average loss for the teacher \cite{he_momentum_2020} or an additional decoder for the student \cite{grill_bootstrap_2020}.  This asymmetry between student and teacher is designed to prevent embedding collapse, but this phenomena is still an active area of research \cite{jain_self-supervised_2022}.

\subsection{Redundancy Reduction}
Like the two contrastive approaches discussed above, redundancy reduction creates augmented views of an image and passes those views through an encoder to generate two embeddings, which it then attempts to align.  But instead of simply aligning embeddings within the latent space, redundancy reduction generates an empirical cross correlation matrix between the two embeddings.  The matrix is NxN, where N is the number of dimensions in the latent space, and each value ranges from -1 (perfect anti-correlation) to 1 (perfect correlation).  Redundancy reduction loss is designed to make the resulting cross correlation matrix resemble the identity matrix.

Redundancy reduction approaches can differ in exactly how that loss is calculated.  Barlow Twins \cite{zbontar_barlow_2021} utilizes two terms: an invariance term and a redundancy reduction term.  The invariance term forces the diagonal of the matrix to values of 1, while the redundancy reduction term forces the non-diagonal elements of the matrix to values of 0.  This shares similar goals to InfoNCE loss, where the invariance term draws embeddings closer together (like positive pairs), and the redundancy reduction term forces embeddings further apart (like negative pairs). However, InfoNCE loss does this across a batch of embeddings, whereas Barlow Twins' loss does so within a given embedding.  VICReg \cite{bardes_vicreg_2022}, builds on Barlow Twins’ intuition and includes three terms: a variance term to prevent collapse, an invariance term based on mean squared error between paired embeddings, and a covariance term to reduce redundancy among feature dimensions.

Redundancy reduction approaches, like distillation SSL, do not require as large of a batch size compared to negative sampling in order to learn effectively \cite{bardes_vicreg_2022}.  Barlow Twins also demonstrated an intriguing ability to more effectively leverage very-high-dimension embeddings: performance increased linearly with regard to the number of dimensions in the latent space.  This trend was observed up to a latent dimensionality of 16K. \cite{zbontar_barlow_2021}.  

\begin{figure}[ht]
    \centering
    \includegraphics[width=0.7\linewidth]{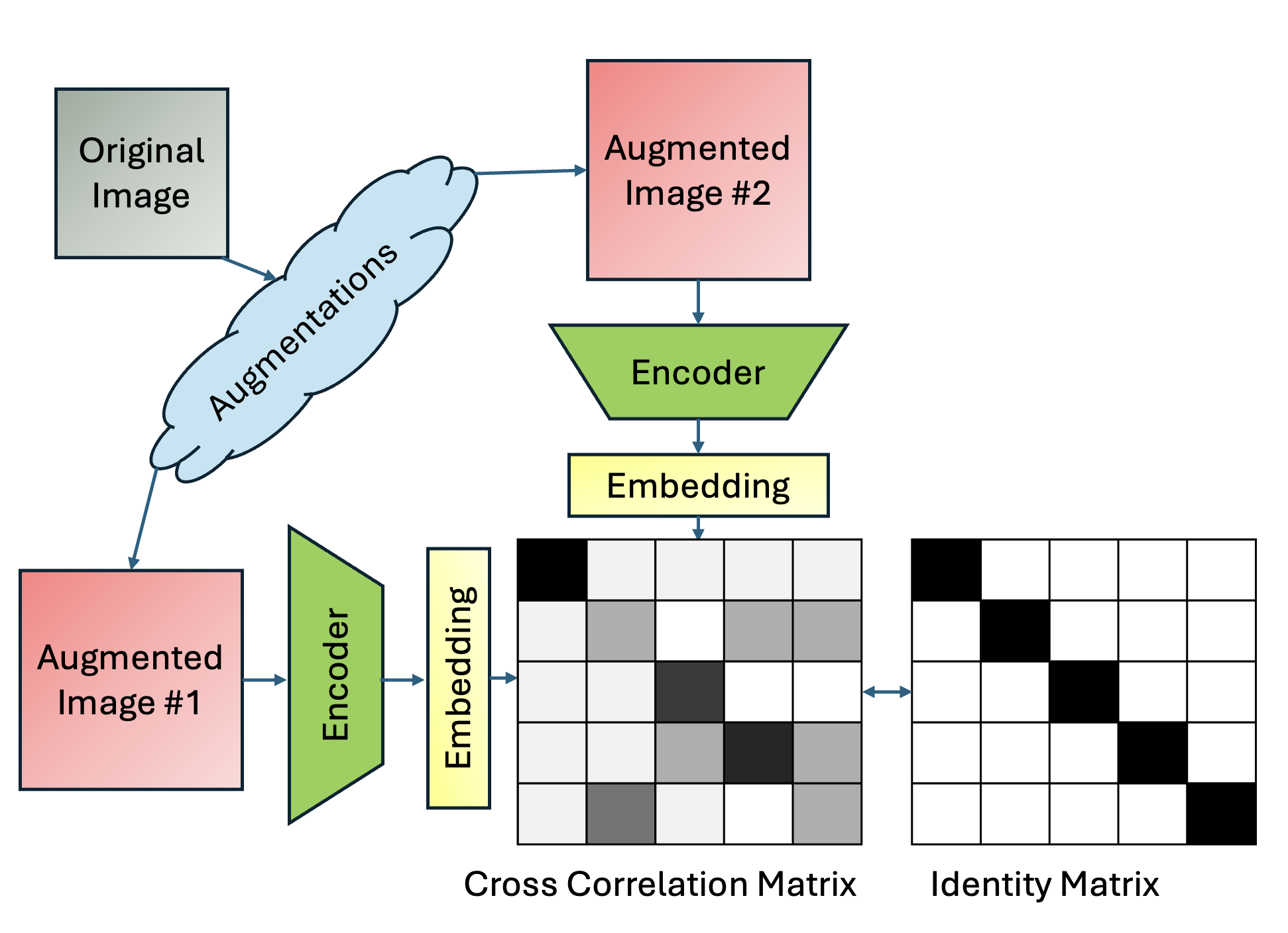}
    \caption{An illustration of a CV foundation model that relies on redundancy reduction.}
    \Description{A diagram that depicts learning via redundancy reduction.  A single image is augmented, and passed through an encoder to generate two encodings.  A cross-correlation matrix is calculated, and then compared to the identity matrix for learning.}
    \label{fig:redundancy_reduction_framework}
\end{figure}

\subsection{Masked Image Modeling}
 Masked Image Modeling (MIM) refers to self-supervised frameworks that reconstructs masked portions of an input image.  MIM is inspired by the Masked Language Modeling (MLM) paradigm introduced in BERT \cite{devlin_bert_2019}. It is worth noting that this is different from autoregressive models such as GPT \cite{radford_language_nodate}, which learn to predict the next token in sequence rather than reconstruct masked inputs. 

 MIM generally entails the following steps:
\begin{enumerate}
    \item Parts of a sample are masked.
    \item The masked sample is run through an encoder to generate embeddings.
    \item A decoder is asked to reconstruct the original sample from those embeddings.
\end{enumerate}
 
 Within MIM, several approaches have emerged beyond the widely used Masked Autoencoder (MAE) framework \cite{he_masked_2021}, including BEiT \cite{bao_beit_2022} (which predicts discrete visual tokens from a pre-trained tokenizer), SimMIM \cite{xie_simmim_2022} (which reconstructs raw pixels of masked patches directly), and iBOT \cite{zhou_ibot_2022} (which predicts masked patch embeddings through self-distillation).  These differ primarily in their masking strategies, prediction targets, and encoder–decoder configurations.
 While the CV literature now spans this diversity of masked modeling techniques, MAE tends to be a popular option.  Research in remote sensing has so far focused mainly on MAE-style methods due to their scalability and efficiency on large unlabeled satellite datasets \cite{he_masked_2021}, while incorporating ideas from other approaches to enhance representation learning. As such, we will tend to put central focus on MAE within the context of this paper. 

MAE masks roughly 75\% of the data to prevent trivial reconstruction from patch context \cite{he_masked_2021}.  The generative nature of the task means that the batch size can be relatively small, making it resource-friendly.  MIM approaches with established frameworks, such as Convolutional Neural Networks (CNNs) or Recursive Neural Networks (RNNs), replace a patch with a mask token, then feed in the whole image, including masked patches, into a network.  But in MAE, the usage of ViTs allows for training pipelines to only feed encoders unmasked patches.  Positional embeddings, in addition to patch embeddings, allows the ViT to avoid processing masked tokens.  This shrinks the size of the necessary input layers and dramatically speeds up training time \cite{he_masked_2021}.   Similarly, utilization of a lightweight decoder also forces the encoder to learn more accurate representations while speeding up runtime \cite{he_masked_2021}.  These optimizations mean that the MAE approach scales remarkably well, resulting in models that can contain billions of parameters \cite{dehghani_scaling_2023}.  Current CV research has identified a gap in the number of parameters between SOTA NLP and CV models.  There are plans to improve the parameter count in CV foundation models even further \cite{dehghani_scaling_2023}.

\begin{figure}[ht]
    \centering
    \includegraphics[width=0.7\linewidth]{images/FM_diagrams/mae.png}
    \caption{CV foundation model framework for masked image modeling}
    \Description{A diagram depicting masked image modeling, where a given image is obscured by a large amount, and then passed through an encoder to generate an embedding.  That embedding is then passed through a decoder which attempts to recreate the original image.}
    \label{fig:mae_diagram}
\end{figure}

\section{Negative Sampling Models in Remote Sensing}
While the underlying principles of negative sampling generally remain the same from CV to RS, RS images often share temporal or geospatial qualities. \cite{jain_self-supervised_2022}.  For example, Sentinel-1 and Sentinel-2 satellites can have images taken of the same location during the same day.  These shared spatiotemporal attributes enables two potential advantages for RS foundation models: 
\begin{enumerate}
    \item Models can leverage imagery that comes from different RS sensors or different temporal acquisitions as naturally occurring augmentations.
    \item It reduces the time required to do hyper-parameter tuning for augmentations.
\end{enumerate}  
The latter often adds significant overhead to training contrastive learning approaches \cite{jain_multimodal_2022}.

\subsection{Temporal Pairing}
Pre-trained on RGB bands from Sentinel-2, Seasonal Contrast (SeCo) \cite{manas_seasonal_2021} borrows heavily from SimCLR’s general framework and outperforms CV models such as MoCo on downstream land cover classification tasks.  This indicates that there is implicit benefit in creating foundation models for the remote sensing domain as opposed to borrowing models directly from the CV field.  Instead of only using hand-crafted image augmentations outlined in SimCLR \cite{chen_simple_2020}, such as color jitter or gaussian blur, SeCo treats images of the same location at different points in time as natural augmentations of one another.  However, SeCo does not simply rely on these natural augmentations.  It also utilizes random cropping, random flipping, and color jitter to prevent the model from learning trivial representations of the data. The resulting model is essentially season- or time-agnostic, meaning it provides a general representation of data regardless of acquisition-time-of-year. This is helpful in seasonally invariant tasks, such as identifying land cover.  However, it may not fit scenarios where retaining seasonal differences in embeddings for downstream tasks is essential, such as projecting forest fire recovery in an area. 

SeCo's approach of contrastive learning between spatially aligned, temporally different RS image pairs has also been explored in Geographically Aware Self Supervised Learning (GASSL) \cite{ayush_geography-aware_2021}.  However, GASSL trains on the Functional Map of the World (FMoW) \cite{christie_functional_2018} dataset, which contains multispectral imagery from the DigitalGlobe constellation \cite{anderson_worldview-2_2012}, and does not utilize any additional artificial augmentations.  In addition to the contrastive task, GASSL also leverages the metadata available in RS imagery for a SSL task.  This approach clusters together images by location and then predicts the latitude and longitude present in the metadata.  The combination of this task coupled with contrastive learning across geographically aligned images results in a framework that attempts to explicitly learn time-agnostic representations of a given place.  As with SeCo, the authors note that this approach learns temporally invariant features, and may struggle with tasks such as change detection \cite{ayush_geography-aware_2021}.  Unlike the approaches below, latitude and longitude are not pre-processed in any way or treated as an additional modality.

\subsection{Geolocation Pairing}
Explicitly utilizing spatial data such as latitude and longitude provides value, but purely introducing them as features can result in models that perform poorly when asked to generalize to locations not within their training data \cite{klemmer_satclip_2024}.  Therefore, a new generation of models dubbed “location encoders” have emerged.  Once trained, a location encoder can take in latitude and longitude and return an embedding of that location which quantifies and summarizes ground conditions present there.  Several models in this generation of location encoders rely on the CLIP framework \cite{radford_learning_2021} and treat ‘location’ as one modality, with a location-specific encoder to use in contrastive learning. This location encoder, in turn, learns the information extracted from the other mode, often an image obtained at that location.

GeoCLIP \cite{cepeda_geoclip_2023} uses images taken at ground level from Flickr for one modality and spatial encodings for the other modality.  GeoCLIP then uses CLIP’s contrastive learning objective to learn effective representations of each modality.  GeoCLIP uses Random Fourier Features (RFF), which have been shown to be highly effective in memorization and image reconstruction \cite{tancik_fourier_2020}, in order to store features learned from images into the location encoder.  These Random Fourier Features allow GeoCLIP to learn high quality spatial encodings that capture information from the ground level photos.

Contrastive Spatial Pretraining (CSP) \cite{mai_csp_2023} also performs contrastive learning between ground level imagery and locations, but it explores a variety of techniques for generating positive and negative pairs that are explicitly tailored to spatially distributed data.  In addition to CLIP’s standard in-batch negative sampling, CSP utilizes random negative location sampling, in which negative pairs for an image are generated by uniformly sampling locations at pre-training time.  In this instance, the positive pair is still the corresponding location for the image.  CSP also explores SimCSE-based sampling \cite{gao_simcse_2022}, in which two location encoders are initialized the same, but utilize different dropout masks.  In this case, the same location is fed through each location encoder, and embeddings that point to the same initial location are considered positive pairs.  All other combinations are considered to be negative pairs.  Each sampling approach forms a component of CSP’s contrastive loss function, resulting in a framework that learns nuanced representations of each modality.  CSP trains on ground-level imagery from the iNaturalist dataset \cite{van_horn_inaturalist_2018}.

SatCLIP \cite{klemmer_satclip_2024} also follows the CLIP approach while utilizing spatial encodings as one of its modalities and optical imagery captured from Sentinel-2 as the other.  SatCLIP does not limit itself purely to RGB data, and rather trains on the 12 bands of data present in Sentinel-2's L2A data product.  SatCLIP does not utilize Random Fourier Features for the spatial encodings, and instead uses a SirenNet \cite{sitzmann_implicit_2020} that relies on spherical harmonics \cite{ruswurm_geographic_2024} to learn effective spatial encodings.

Geo-Aligned Implicit Representations (GAIR) \cite{liu_gair_2025} takes this a step further by utilizing location as a way to align multispectral satellite images from Sentinel-2 and ground-level RGB images from the Global Streetscapes dataset \cite{hou_global_2024}.  As with SatCLIP, GAIR trains on Sentinel-2's L2A data product.  Ground-level imagery can provide a highly informative mode to train satellite imagery against.  However, a single satellite image's scope is so large that hundreds or thousands of ground-level images can correspond to a single satellite image.  In order to address this issue from a contrastive learning standpoint, GAIR learns Implicit Neural Representations (INR) \cite{xu_signal_2022} of a Sentinel-2 image.  INR learns a continuous function for spatial coordinates to corresponding signals.  These INR encodings of Sentinel-2 imagery are then contrasted against location encodings and ground level vision encodings.  GAIR uses RFF \cite{tancik_fourier_2020} for location encodings and a standard ViT for ground-level image encodings.

\subsection{Multisensor Pairing}
Multi-modal SimCLR \cite{jain_multimodal_2022} treats images of the same location from different sensors as naturally occurring augmentations.  This approach has two encoders: one trained on RGB bands of  Sentinel-2, and one trained on VV and VH polarizations of SAR images from Sentinel-1.  As with SeCo, this approach does not rely entirely on natural augmentations.  It adds random cropping, flipping, color jitter/drop, and gaussian blur, all augmentations that have proven useful in other remote sensing models \cite{jain_self-supervised_2022}.  Interestingly, fine-tuning Multi-modal SimCLR on downstream tasks with a dataset of only one modality, either Sentinel-1 or Sentinel-2, yields better results compared to other methods that train only on the one sensor \cite{jain_multimodal_2022}.  This indicates that leveraging the additional modalities (i.e., sensors) present in remote sensing for SSL has the potential to result in better representations of a single modality in the modality-specific encoder.

\subsection{Image-Text Pairing}
CLIP’s original design used text and images as its two modalities, and RemoteCLIP \cite{liu_remoteclip_2024} goes back to these roots. RemoteCLIP takes in text as its one modality, leaning on image captions to generate natural labels.   Other models' requirements for additional fine-tuning, as opposed to zero-shot utilization, motivates this model’s development.  Including text as one of the modalities results in good zero-shot performance downstream \cite{zhu_foundations_2024}.  This model treats RGB images taken from Sentinel-2, drones, and other aerial platforms as one modality.  This variety in the sources for the image modality is designed to result in a more robust encoder that will be resolution agnostic.  RemoteCLIP's reported underperformance in zero-shot image classification tasks indicates that there is additional work to be done in that regard.  

For the text modality, RemoteCLIP relies on image captions already present in the human-annotated satellite imagery datasets.  Where no captions are paired with an image, a preprocessing stage generates appropriate text annotations.  This preprocessing stage performs standard object detection on the image, then uses a script to programmatically generate a corresponding sentence.  That sentence is then used as the positive text annotation. Generating captions via script will lack natural variance that is present in a typical corpus, and generating these captions via NLP processes such as ChatGPT will inherently introduce extrinsic noise and potential bias into the dataset.  Nevertheless, RemoteCLIP outperforms SOTA models like CLIP and DINOv2 in zero-shot evaluation of object counting in the Remote-Count dataset \cite{gao_counting_2021}, and few-shot classification of 12 RS benchmarks, including EuroSat \cite{helber_eurosat_2019}, RSC11 \cite{zhao_feature_2016}, and RESISC45 \cite{cheng_remote_2017}.  This performance indicates that text as a modality has definite value in the remote sensing domain.

Given multi-modal foundation models' origins as VLMs, there are a variety of other RS foundation models that leverage text as a modality in order to enable few-shot learning capabilities.  These models have been surveyed by \cite{zhou_towards_2024}.

\begin{figure}[ht]
    \centering
    \includegraphics[width=0.9\linewidth]{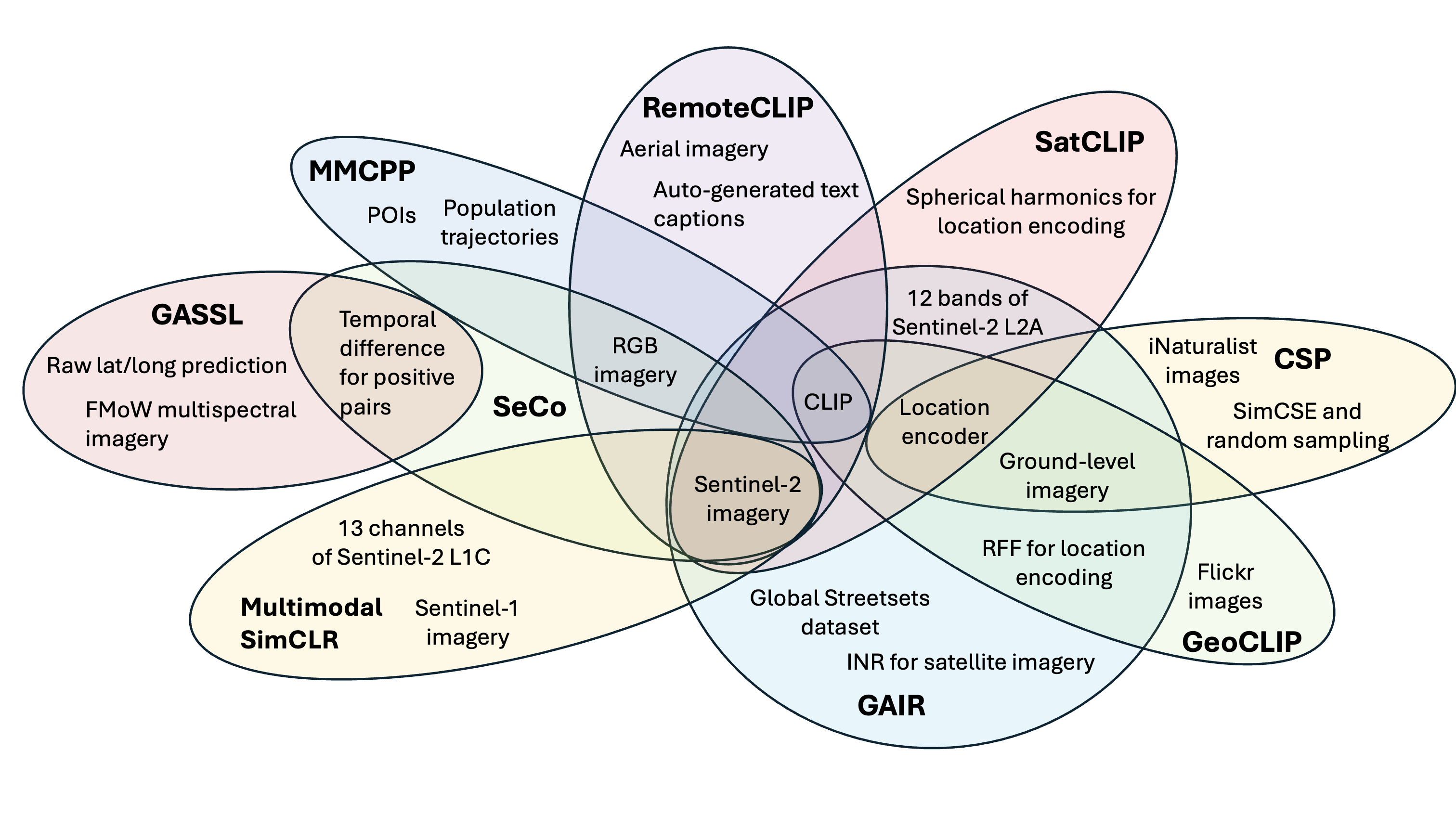}
    \caption{An illustration of shared and unique concepts among RS foundation models which utilize contrastive learning with negative sampling for their SSL task.}
    \Description{A sort of Venn diagram where a variety of RS models that utilize contrastive learning with negative sampling are compared with one another to highlight similarities and differences.}
    \label{fig:contrastive_models_breakdown}
\end{figure}

\section{Distillation Networks in Remote Sensing}
Given the requirement for handling a sizable amount of remotely sensed data, efficiency-oriented SSL approaches like distillation have become popular for remote sensing foundation models.  

\subsection{Sensor-Agnostic Encoder}
One successful example of a distillation network in the RS field is RS-BYOL \cite{jain_multimodal_2022}, which borrows BYOL’s framework and uses the multi-modal aspect for multi-sensor remote sensing data.  Just as with negative sampling approaches in RS, different sensors’ imaging of the same location are leveraged as natural augmentations of the same image.  One modality is drawn from Sentinel-2 optical, whereas the other is drawn from Sentinel-1 SAR.  Similarly, additional augmentations of random crops, rotations, color jitter, and grayscale are applied to prevent the model from learning trivial representations of the data.  RS-BYOL utilizes 10 bands of data in Sentinel-2 with 10-20 m resolution and the VV and VH polarizations of SAR present in Sentinel-1.  RS-BOYL only trains a single encoder at a time and chooses 3 random bands of Sentinel-2 and 3 random polarizations from Sentinel-1 to make up the two image views.  They perform distillation with view 1 passing through the student encoder and view 2 passing through the teacher encoder, then also perform distillation with the views passing through the opposite encoders.  The resulting framework represents an early attempt to learn sensor-invariant features.  

The model outperforms negative sampling approaches such as SeCo in linear probing when all bands are present from Sentinel-2, indicating that there is inherent value in incorporating multispectral bands from Sentinel-2 in addition to the RGB bands.  However, SeCo outperforms RS-BYOL with linear probing when only the RGB bands are present, indicating that negative sampling approaches may still have a performance edge in remote sensing.  The lighter resource requirements for distillation networks reflects a tradeoff between model performance and hardware requirements.  RS-BYOL represents a somewhat straightforward adaptation of CV distillation SSL techniques in RS, but other approaches make more dramatic changes.

DINO-MM \cite{wang_dino_mm_2022} is another RS foundation model leveraging distillation as its SSL objective, although this model is built off of DINO's \cite{caron_emerging_2021} framework instead.  DINO-MM trains on BigEarthNet-MM \cite{sumbul_bigearthnet_2019}, a dataset containing roughly half a million pairs of Sentinel-2 and Sentinel-1  images at the same spatiotemporal scenes.  DINO-MM represents one of the first RS foundation models to explore the use of ViTs \cite{dosovitskiy_image_2021} as a backbone.  As such, its student and teacher encoders can handle flexible input shapes.  DINO-MM takes full advantage of this, stacking SAR and MS channels on top of one another to create the 'original image'.  DINO-MM applies the standard augmentations present in DINO, but also introduces a RandomSensorDrop augmentation that results in one of three outcomes:
\begin{enumerate}
    \item All SAR channels are dropped
    \item All MS channels are dropped
    \item No channels are dropped
\end{enumerate}

This results in an encoder that is capable of leveraging whatever satellite imagery is available when creating embeddings.  In theory, it also means that this model learns intra- and inter-modal representations of the data.  However, this framework's flexibility does not necessarily guarantee performance when all modes of data are present.  Reported ablation studies show that downstream assessment on BigEarthNet's classification task perform well with Sentinel-2 data, but do not actually improve when given additional Sentinel-1 information.  In the case of top-1 benchmarking, the model's performance actually decreased slightly while fine-tuning on Sentinel-1 and Sentinel-2 imagery (when compared to purely Sentinel-2 fine-tuning).  It is possible that benchmarking on tasks where SAR data is more relevant, such as Sen1Floods11 \cite{bonafilia_sen1floods11_2020}, might yield more favorable results with multiple modes of data present.

\subsection{Multiple Contrastive Tasks}
SkySense \cite{guo_skysense_2024} represents a significant step in RS foundation model size, with 2.09 billion trainable parameters.  Aside from model size, SkySense performs distillation SSL at multiple stages of its training pipeline, resulting in a model capable of learning effective high level and low level features.  SkySense trains on three modalities: time series of Sentinel-1, time series of Sentinel-2, and high resolution RGB imagery from Worldview 3 and 4.  For each mode, there is a teacher and student encoder.  SkySense performs two sets of random augmentations per image, then sends the different views to the student and teacher.  It performs multi-modal contrastive loss between the embeddings from the student encoders and then performs a number of contrastive learning tasks between student and teacher embeddings, namely:
\begin{itemize}
    \item pixel-level contrastive loss
    \item image-level contrastive loss via average pooling of pixels
    \item object-level contrastive loss by using the Sinkhorn-Knopp algorithm \cite{knight_sinkhornknopp_2008} to perform unspervised clustering on pixel-level features
\end{itemize}

After these contrastive tasks, the embeddings go through a multi-modal temporal fusion encoder at both the student and teacher level.  This results in a fused embedding.  Sinkhorn-Knopp is applied again, but this time, the resulting clusters are compared to geographic regions in order to learn geo-context.  Then the fused embeddings go through the same contrastive learning tasks described in the paragraph above.  The resulting model was reported to achieve SOTA performance on a plethora of downstream tasks at the time of publication in 2024.  These tasks included:
\begin{itemize}
    \item Segmentation tasks, such as Potsdam \cite{sherrah_fully_2016}
    \item Object detection tasks, such as DIOR\cite{li_object_2020}
    \item Change detection tasks, such as OSCD\cite{daudt_urban_2018}
    \item Classification tasks, such as MillionAID\cite{long_creating_2021}
\end{itemize}.

\begin{figure}[ht]
    \centering
    \includegraphics[width=0.7\linewidth]{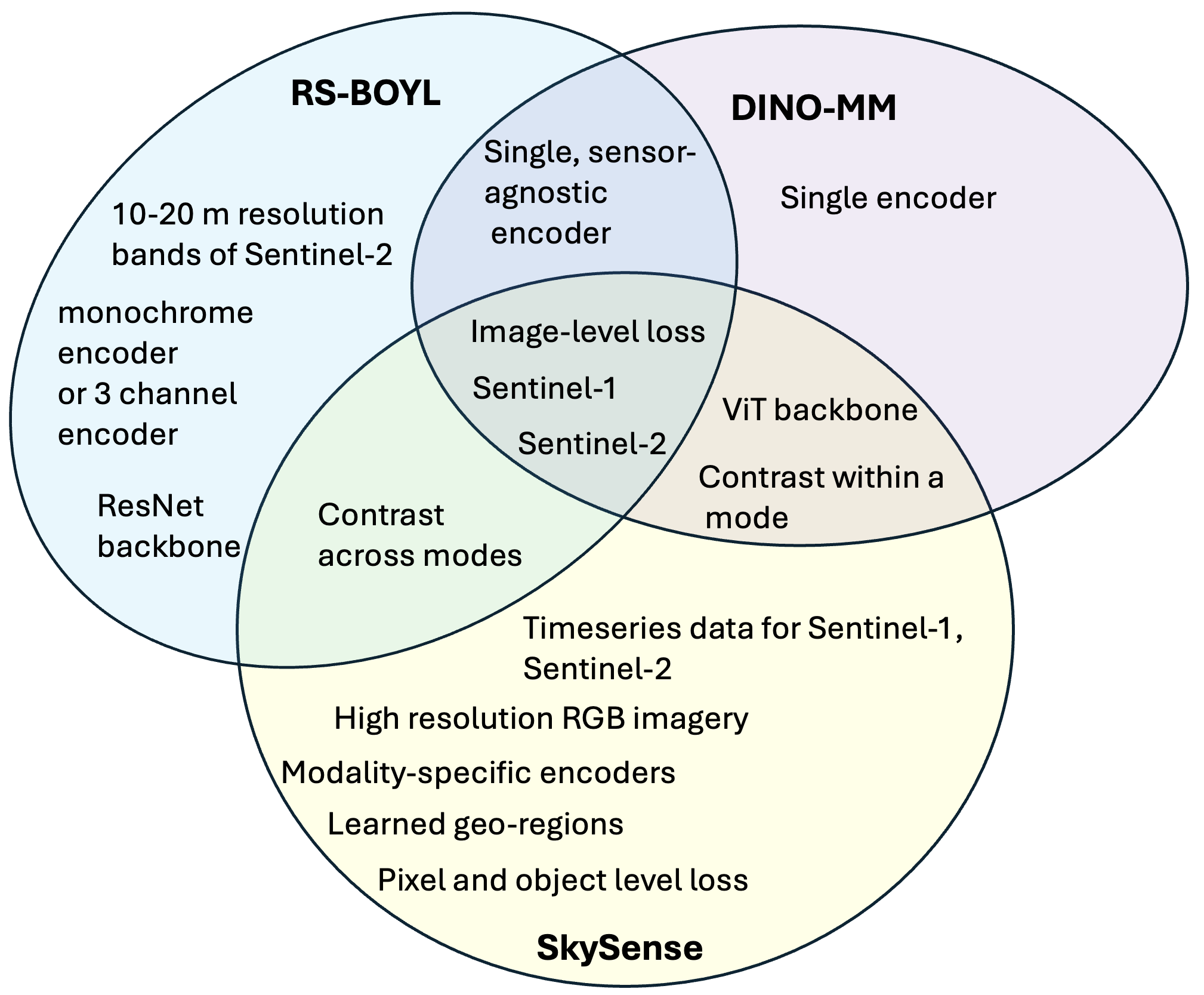}
    \caption{An illustration of shared and unique concepts among RS foundation models which utilize knowledge distillation for their SSL task.}
    \Description{A sort of Venn diagram where a variety of RS models that utilize distillation SSL are compared with one another to highlight similarities and differences.}
    \label{fig:distillation_models_breakdown}
\end{figure}

\section{Redundancy Reduction in Remote Sensing}
Decoupling Common and Unique Representations (DeCUR) for Multimodal SSL \cite{wang_decoupling_2024} builds on the Barlow Twins redundancy reduction approach, but modifies it to handle multiple modes of data.  DeCUR explores several multi-modal training datasets.  In order to explore the efficacy of their multi-modal training approach, DeCUR trains one model per each of the following datasets:
\begin{itemize}
    \item SAR/optical imagery from SSL4EO \cite{wang_ssl4eo-s12_2023}
    \item RGB imagery/DEM data from GeoNRW \cite{baier_geonrw_2020}
    \item RGB imagery/depth data from SUN-GRBD \cite{song_sun_2015}
\end{itemize}

DeCUR's key distinction from other multi-modal approaches we have discussed lies in its desire to learn inter-modal representations, but preserve data unique to a given modality.  It does so by setting aside a percentage of the latent space for shared representations and leaving the other percentage of the latent space for modality specific information.  When the Barlow Twin's redundancy reduction loss is applied from one mode to another, DeCUR only calculates this loss for shared dimensions of the latent space.  DeCUR also calculates redundancy reduction loss within a given mode itself, and does so for all of the dimensions within the latent space.  This strategy enables DeCUR to align modalities without losing modality-specific information and obtained SOTA performance on BigEarthNet-MM \cite{sumbul_bigearthnet_2019} classification tasks at the time of publication in 2024.

One potential downside of this approach is the calculation of the shared latent space dimensions.  The authors use a light grid search in each dataset to determine how much of the latent space should be used for learning shared representations.  This percentage tends to hover around 75-80\% in the datasets above, but calculating it becomes costly with larger datasets.  The authors also found no significant drop in performance when they dropped the shared dimensions to as low as 50\%.  This suggests a degree of sparsity in the shared latent space and an opportunity to refine how the shared dimension percentage is calculated, or ways to even better leverage all dimensions.

\section{Masked Image Modeling in Remote Sensing}
While MIM is a powerful approach in CV, there is significant room to optimize performance for remotely sensed imagery.  Below are some of the adjustments that have been made to MIM in order to leverage it more effectively in the remote sensing domain.  Most models discussed follow the MAE masking and decoder paradigm, with a few noted exceptions.

\subsection{Masking Approach}
The simple CV approach of masking individual patches of an image can work with RS imagery, but there is room for improvement \cite{sun_ringmo_2023}.  Compared to natural images in CV, satellite and aerial images in RS often capture small surface features, dense objects, complex backgrounds, and a range of possible object orientations.  Similarly, natural images tend to follow expected proportions with well-defined distributions, whereas the same cannot be said for RS images \cite{rolf_mission_2024}.   Sun’s Remote Sensing Foundation Model (RingMo) introduces PIMask \cite{sun_ringmo_2023} to account for these inherent differences, as masking an entire patch of a RS image might lose information that is unique to that given patch.  Therefore, some pixels in each masked patch are randomly preserved, and multilayer convolution is applied to each patch.  Compared to traditional CV methods, where masking 50\% of an image would completely mask 50\% of the patches, masking 50\% of an image with PIMask would partially mask a higher percentage of patches.  The overall number of pixels masked in the image would still be 50\%, but the spatial distribution is different.  RingMo is officially classified as a MIM approach, as the masking strategy does not allow for the encoder to only be fed unmasked patches.

MAE in CV finds that masking high percentages of the image (roughly 75\%) results in more accurate representations being learned \cite{he_masked_2021}, and that carries over to the current implementations of MAE in remote sensing as well \cite{sun_ringmo_2023}.  This parallels with the findings from negative sampling approaches that making the pretext task with contrastive objective more difficult via multiple augmentations results in the model learning better embeddings.  RingMo’s embeddings outperform embeddings generated with comparable MIM frameworks and standard CV patch masking, underscoring the importance of improving the masking procedure for satellite imagery.

Spatial-Spectral MAE (S2MAE) \cite{li_s2mae_2024} takes this a step further, as they find that masking 90\% of satellite imagery yields the best results in their novel framework.  S2MAE trains on multispectral imagery from FMoW \cite{christie_functional_2018} and BigEarthNet \cite{sumbul_bigearthnet_2019} datasets.  Unlike RingMo and other image RGB masking strategies which mask out the same pixel across all channels, S2MAE employs independent 3D masking, which masks out different pixels in each channel.  3D masking was introduced for videos and appears to learn environmental details well, but struggles to track movement from frame to frame \cite{feichtenhofer_masked_2022}.  While this is poorly suited for video analysis, it is ideal for RS applications, as seen in S2MAE's SOTA performance on the Onera Satellite Change Detection (OSCD) \cite{daudt_urban_2018} and EuroSat \cite{helber_eurosat_2019} benchmarks at the time of publication in 2024.  S2MAE also finds that shallow decoders are poorly suited for MAE with multispectral imagery, a notable departure from established CV practices \cite{he_masked_2021}.

SatMAE \cite{cong_satmae_2023} also experiments with different masking approaches.  It utilizes consistent masking, where the masked regions are spatially-consistent across all images of a given location, and independent masking, where the location of the patches can change from image to image of a given location. Results indicate that independent masking can trivialize the task by leveraging spectral redundancy.  In other words, the model might be able to reconstruct a given masked patch by referring to an unobstructed view of that patch at a given time. A random cropping augmentation is applied to address this and ensure the model learns robust embeddings.

SpectralGPT \cite{hong_spectralgpt_2024} experiments with a unique 3D-cube masking strategy.  Training on 12 bands of Sentinel-2 data from FMoW \cite{christie_functional_2018} and BigEarthNet \cite{sumbul_bigearthnet_2019}, SpectralGPT's approach cannot be described as either independent or consistent masking.  Instead of tokenizing channel by channel, SpectralGPT creates non-overlapping tokens which span 3 channels.  Independent masking is then applied to each layer of tokens.  This results in a blend of consistent and independent masking that is informed by spectral resolution.  SpectralGPT also uses multiple loss terms in their approach, generating loss based on token-to-token reconstruction, but also spectral signature-to-spectral signature reconstruction.  Similar to S2MAE \cite{li_s2mae_2024}, SpectralGPT finds that it performs best with a masking rate of 90\% and a deeper decoder.

\subsection{Temporal Encodings}
SatMAE \cite{cong_satmae_2023} leans on MAE’s utilization of a ViT as the backbone, but encodes the hour, month, and year as temporal encodings.  SatMAE then concatenates these temporal encodings to the standard positional encodings that are innate to the functioning of a ViT.  As with SeCo, the inclusion of temporal information in self-supervised learning for RS foundation models can have substantial performance improvements, but it also means that the embeddings learned are agnostic to temporal variability.  Thus, the embeddings' applicability to temporally sensitive downstream tasks is limited.  Still, this ability to encapsulate temporal data highlights the flexibility of the MAE approach in remote sensing.

Prithvi \cite{jakubik_foundation_2023} trains on the Harmonized Landsat Sentinel-2 (HLS) \cite{claverie_harmonized_2018} dataset, which contains multispectral images captured by Landsat 8/9 and Sentinel-2.  Given the differing revisit times of the two satellites, HLS is a satellite imagery dataset with a high temporal resolution.  Prithvi leverages this higher temporal resolution to arrange these 2D images in a time series and updates the traditional MAE positional and patch embeddings to be 3-dimensional, where the third dimension pertains to time.  When masking patches, they specifically apply 3D convolutions to avoid losing information specific to any one given patch, similar to RingMo’s \cite{sun_ringmo_2023} approach.  Unlike RingMo, Prithvi fully masks patches.  Prithvi also utilizes a landscape stratified sampling strategy to avoid bias towards more common ecosystems or landscapes.

Presto \cite{tseng_lightweight_2024} takes this approach even further, constructing a pixel-timeseries training dataset that consists of input from multiple types of data, including Sentinel-2 RGB, Sentinel-1 SAR, DEM, and the Dynamic World \cite{brown_dynamic_2022} dataset.  Where other foundation models typically operate on patches or convolutions of pixels, Presto operates explicitly on pixels.  This choice is more costly from a computational standpoint, but theoretically results in high-resolution representations.  Dynamic World contains land cover labels for 10m resolution Sentinel-2 imagery.  The framework then employs a variety of different strategies for training.  It masks channels, blocks of timesteps, single timesteps, and/or random pixels for reconstruction by a downstream encoder.  The resulting encoder performs well on feature extraction, fine-tuning, and time series forecasting.  The variety of masking strategies ensures that Presto can successfully process time series datasets with missing channels, a variety of temporal resolutions, and only a small subset of timesteps.  However, performance on tasks that are temporally agnostic and rely on a single image lags behind SOTA models.  The authors note that this is possibly due to the model’s inability to process images with a spatial resolution higher than 10m.

\subsection{Spatial Resolution}
Foundation models in CV, and most foundation models in RS by extension, process images in local image coordinates.  This means that the learned representations are dependent on the spatial resolution of the satellite sensor.  ScaleMAE \cite{reed_scale-mae_2023} is designed to address this issue by scaling the image coordinate system such that it matches the actual spatial scale of objects, ensuring that patch size, as measured in geospatial units, is the same across various images.  This is done via introducing a Ground Sample Distance (GSD) positional encoding, which then informs the ViT of both the geospatial scale and image position of the patch. Put differently, ScaleMAE addresses relative spatial resolution, but does not incorporate absolute geolocation in its training.  As with SatMAE, ViT’s framework lends itself well to introducing the concept of spatial resolution to the model.  Dovetailing with the incorporation of spatial resolution, ScaleMAE’s decoder produces both high and low resolution images in order to capture both low-frequency features, such as mountains and rivers, and more fine-grained high-frequency features, such as vegetation.  It does so by utilizing a Laplacian-pyramid decoder \cite{burt_laplacian_1983} instead of the standard transformer decoder.

Where ScaleMAE focuses on understanding spatial resolution of the overall image, USat \cite{irvin_usat_2023} chooses to focus on spatial resolution across image channels.  Different bands of the same sensor can have dramatically different spatial resolutions.  For instance, in the case of Sentinel-2, bands (i.e., input channels) range from 10m resolution to 60m resolution.  Many approaches up- or down- sample channels to yield a consistent spatial resolution, and some simply drop low resolution channels altogether \cite{jain_self-supervised_2022}.  This ensures that patch sizes remain consistent, but may lose resolution specific information.  USat trains on Sentinel-2 and NAIP, and adjusts the number of patches per channel depending on the spatial resolution.  High resolution channels have more patches to capture specific high resolution information.  Lower resolution channels have fewer patches, but the total number of patches must be evenly divisible by the number of high resolution patches.  This enables channels with different resolutions to cleanly map to one another.  Each channel will be masked by the same percentage. More patches are masked in high resolution channels to preserve the percentage of the image being masked.  Then, patches are aggregated via spectral pooling, and a standard decoder is used to reconstruct the original image.  This approach effectively extracts high-level and low-level features from differing resolutions, but does have some drawbacks.  Namely, the assumption that different sensors will have resolutions that are evenly divisible will not always be true.  The authors also note that the approach becomes costly to train with multiple different sensors.

SatMAE++ \cite{noman_rethinking_2024} takes inspiration from ScaleMAE's approach and seeks to introduce spatial resolution awareness to models that train on non-RGB modalities, training on Sentinel-2 imagery in FMoW \cite{christie_functional_2018}.  They go about doing so with a more straightforward approach than ScaleMAE, doing away with GSD encodings and Laplacian Pyramid decoders.  Instead, SatMAE++ has a multi-resolution reconstruction objective.  An image is upsampled 2x and 4x, masking occurs, and then the same decoder is responsible for reconstructing different resolution images based on the corresponding masked input.  For the original image, standard mean squared error is applied.  For the upsampled blocks, L1 loss is used.

\begin{figure}[ht]
    \centering
    \includegraphics[width=0.9\linewidth]{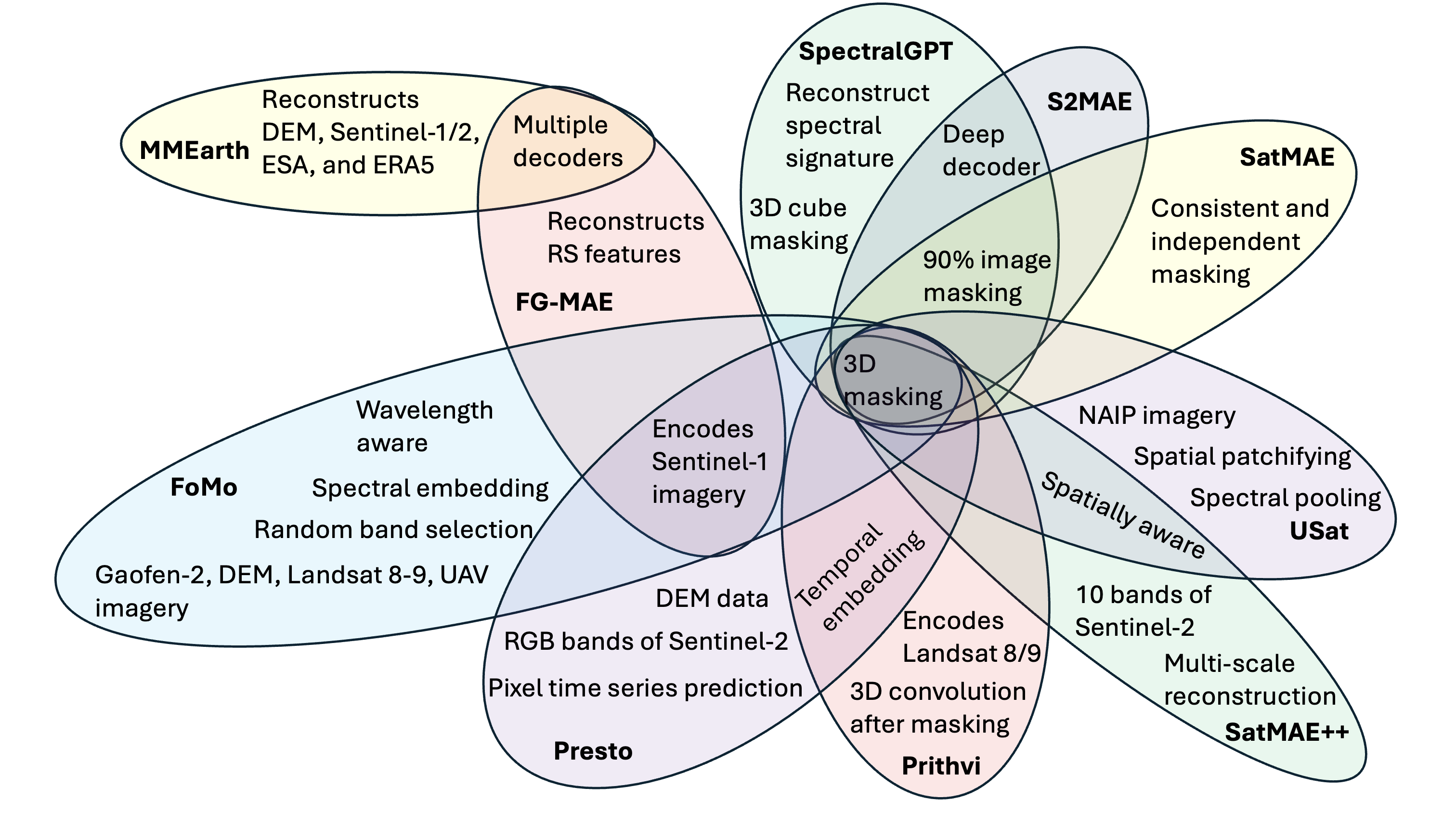}
    \caption{An illustration of shared and unique concepts among RS foundation models which utilize masked auto-encoding for their SSL task.  These models all encode multispectral imagery from Sentinel-2 and potentially some other multi-sensor data.}
    \Description{A sort of Venn diagram where a variety of remote sensing masked auto-encoding foundation models are compared and contrasted to highlight their similarities and differences.  The models in question train on data acquired from multiple sensors or contain imagery captured by a multispectral sensor.}
    \label{fig:multispectral_mim}
\end{figure}

\subsection{Multiple Decoders}
MAE is originally designed to take in a single modality, which can make it challenging to utilize the different sensors that are available in remote sensing.  By comparison, contrastive learning models elegantly use different modes of data available by creating an encoder per modality.  MMEarth \cite{nedungadi_mmearth_2024} explores a potential multi-modal MAE approach by pairing an encoder for one modality with many modality or task-specific decoders.   MMEarth takes in multispectral data from Sentinel-2 that contains all 13 bands, masks that data, then utilizes an encoder to generate embeddings.  Typical MAE would pair this with a single decoder designed to reconstruct the original optical image.  Instead, MMEarth pairs the optical encoder with 12 separate decoders.  Six of these decoders are focused on pixel level tasks, such as reconstructing a corresponding optical image, a corresponding SAR image, or a corresponding Dynamic Earth image.  The other six decoders are focused on image level tasks, such as classifying the corresponding biome or month.  

The loss for all of these decoders is back propagated into the encoder, resulting in representations of optical imagery that capture data from other modes.  While this approach learns refined optical representations, contrastive learning approaches of a similar manner would also learn representations for the other modes of data.  Depending on the downstream modalities available, this may be a significant limitation for MMEarth’s effectiveness.

Feature Guided MAE (FG-MAE) \cite{wang_feature_2023} also explores the utilization of multiple decoders for MAE, but does so by tasking the model to reconstruct specific features of interest in the remote sensing world.  MAE's focus on pixel-level details can limit the model's capability in understanding satellite imagery.  FG-MAE explores a variety of reconstruction objectives, including Histograms of Oriented Gradients (HOG) \cite{dalal_histograms_2005},  Normalized Difference Indices (NDI) such as Normalized Difference Vegetation Index (NDVI) \cite{huang_commentary_2021}, CannyEdge \cite{canny_computational_1986}, and Scale-Invariant Feature Transform (SIFT) \cite{lowe_distinctive_2004}.  FG-MAE maintains two separate models: one for optical imagery and one for SAR imagery.  In EuroSat \cite{helber_eurosat_2019} and BigEarthNet \cite{sumbul_bigearthnet_2019} classification tasks, FG-MAE's optical model demonstrates comparable or better results to standard MAE when only reconstructing a single one of these RS features.  However, performing reconstruction on both HOG and NDI results in a substantial performance boost compared to standard MAE.  FG-MAE's SAR model finds that simply reconstructing HOG yields the best results, as it tends to work well with the noisier Sentinel-1 imagery.

\subsection{Orientation Awareness}
In natural imagery, the orientation of objects is of importance; if an object is upside-down, that is semantically different compared to the same object right side up.  By comparison, the same object can appear in satellite imagery in a variety of different orientations, and those orientations have little to no semantic differences.  Leveraging the standard MAE approach from CV treats these orientations differently, which is not desirable for remotely sensed images.  

Masked Angle-Aware Auto-Encoding (M3AE) \cite{li_masked_2024} emerges to address this particular problem.  M3AE trains on the MillionAID dataset \cite{long_creating_2021}, an RGB dataset derived from Google Earth imagery.  M3AE introduces a scaling, center crop augmentation to every image, resulting in an augmented image where an object in the center will be rotated into a different orientation from the original.  This augmented image is masked then fed into an encoder-decoder framework where the decoder attempts to reconstruct the original image.  In order to ensure that the model pays attention to the re-oriented object, the encoder is fed positional embeddings that include a third dimension to track the angle of a given patch.  M3AE also introduces an optimal transport reconstruction loss to automatically assign similar image patches for each rotated crop patch.  This ensures that the rotation is taken into effect from a loss standpoint.

Rotated Varied-Size Attention (RVSA) \cite{wang_advancing_2023} also develops a novel approach to developing orientation-agnostic representations of RS imagery, but does so through ViT self-attention compared to image augmentation.  Trained on the MillionAID dataset \cite{long_creating_2021}, RVsA shifts away from the standard multi-head self-attention (MHSA) present in traditional CV ViTs, as full attention scales quadratically to image size.  Given the potential high-resolution nature of RS imagery, RVSA leverages a varied-size window-based MHSA (VSA).  In VSA, images are partitioned into non-overlapping windows, then have MHSA applied \cite{zhang_vsa_2022}.  In order to learn orientation, RVSA randomly rotates and re-sizes these windows before performing self-attention within a ViT.  With these self-attention tweaks in effect, standard MAE is employed for reconstruction masked MillionAID images.  The result is a framework capable of swiftly handling high resolution RS imagery that learns orientation-agnostic representations of RS imagery.

While orientation matters a great deal for object recognition in RS imagery, the M3AE authors notes that this applies predominantly to man-made objects \cite{li_masked_2024}.  These representations are likely better suited for urban or developed environments, and development of a training dataset specifically focusing on man-made objects would likely be useful for model fine-tuning.

\begin{figure}[ht]
    \centering
    \includegraphics[width=0.9\linewidth]{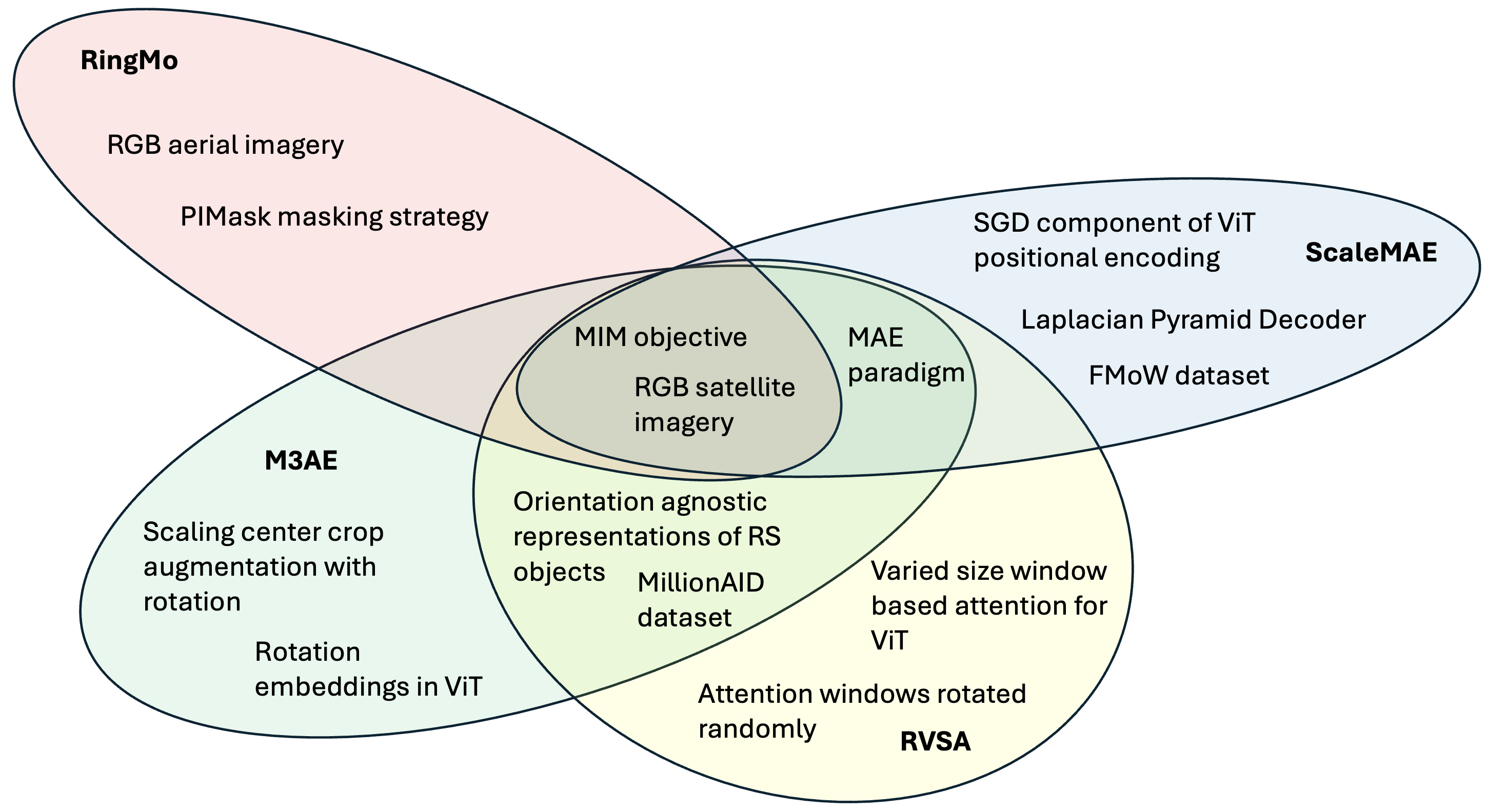}
    \caption{An illustration of shared and unique concepts among RS foundation models which utilize masked image modeling for their SSL task and all train only on RGB images.}
    \Description{A sort of Venn diagram where a variety of remote sensing masked image modeling foundation models are compared and contrasted to highlight their similarities and differences.  The models in question all train only on RGB imagery.}
    \label{fig:mae_rgb}
\end{figure}

\subsection{Wavelength Awareness}
The variety of different satellite sensors available means that even if two sensors monitor the same rough wavelengths, their spectral resolution or means of acquisition might be very different.  For example, Landsat-7 and Sentinel-2 are both considered multispectral and monitor visible to SWIR wavelengths, but Landsat-7 splits that into 8 distinct bands, compared to Sentinel-2's 13 bands.  As a result, a foundation model trained on Sentinel-2 data would not be well-suited to work with Landsat-7 data out-of-the-box, and vice-versa.

Foundation Models for Forest Monitoring (FoMo) \cite{bountos_fomo_2025} trains on four satellite-based datasets and four aerial-based dataset, with spatial resolutions ranging from less than 5cm to 60m.  The datasets include MS imagery from Landsat 8-9, Sentinel-2, and Gaofen-2.  They also include SAR imagery from Sentinel-1, DEM data, and both RGB and MS imagery from high-res UAV sensors.  The resulting training dataset is rich in spatial and spectral diversity.  FoMo's flexible training strategy groups together all of the images of the same spatiotemporal scene, then randomly selects bands to use in that particular sample.  Each band is independently tokenized to avoid spatial resolution issues.  FoMo then incorporates standard positional embeddings and introduces spectral embeddings to denote what band each channel belongs to.  This setup allows for learned representations across bands and within bands.  Tokens are masked, passed through a ViT encoder, and then go through a standard decoder for reconstruction.  Due to the random band selection process for generating samples, batches can potentially be very different.  Gradient accumulation across batches is used to 'smooth' the learning rate.  

FoMo's architecture does not boast the ability to impute information for untrained on wavelengths the way DOFA might.  However, the diversity of the training data and elegant approach yield a model that is effective at leveraging data from a wide swathe of sensors.  The authors do note that independent channel tokenization can also lead to potential scalability issues.  The amount of tokens per sample increases linearly with the number of spectral bands, resulting in the complexity of the self-attention mechanism increasing quadratically.

\section{Combined Approaches}
Several RS foundation models cannot in good conscience be described as falling into any of the above approaches.  Instead, they leverage multiple objectives that span the  approaches discussed above.

Contrastive Radar-Optical Masked Auto-encoders (CROMA) \cite{fuller_croma_2023} is one such model.  CROMA trains on Sentinel-1 and Sentinel-2 imagery and trains an encoder for each sensor.  However, CROMA takes inspiration from the Fast Language Image-Pretraining (FLIP) framework \cite{jia_scaling_2021}, which independently masks each mode of data before feeding them into encoders and performing contrastive learning in the standard CLIP \cite{radford_learning_2021} setup.  Masking images before encoding them results in a significantly less memory intensive process than encoding the raw image when vision transformers are used.  Therefore, the FLIP framework enables contrastive training with a much larger batch size with the same or less memory footprint.  This results in faster convergence.  CROMA implements this approach with remotely sensed images, contrastively learning between masked Sentinel-1 and Sentinel-2 images.  CROMA also introduces a third encoder that takes in concatenated Sentinel-1 and Sentinel-2 encodings to generate multi-sensor encodings.  These fused encodings are fed into a multi-sensor decoder responsible for constructing the initial Sentinel-1 and Sentinel-2 images.  In this manner, CROMA is able to introduce contrastive and reconstruction loss terms into their framework.  This results in a model that effectively learns uni-sensor representations for Sentinel-1 and Sentinel-2 and multi-sensor representations of both Sentinel-1 and Sentinel-2.  CROMA's framework is benchmarked against BigEarthNet \cite{sumbul_bigearthnet_2019} and FMoW \cite{christie_functional_2018} classification tasks, demonstrating SOTA performance with images from a single sensor and multi-sensed images at the time of publication in 2023.  BigEarthNet contains roughly half a million Sentinel-2 images where each patch can have multiple land-cover classes.  CROMA also utilizes modified Attention with Linear Biases (ALiBi) positional embeddings \cite{press_train_2022} for their ViT encoders.  These embeddings provide an advantage over standard sinusoidal positional embeddings by biasing the embeddings based on Euclidean distance between key-query pairs, thus imbuing the embeddings with a degree of spatial awareness.

OmniSat \cite{astruc_omnisat_2024} also leverages an encoder per modality, but does so for three separate modes of data: \begin{enumerate}
    \item timeseries of multispectral imagery
    \item timeseries of SAR imagery
    \item high resolution UAV imagery that features RGB bands, a NIR band, and DEM data
\end{enumerate}  
After patchifying and encoding the data through their respective encoders, OmniSat then performs contrastive loss between the patches of each modality, where each patch has M-1 positive pairs (M representing the number of modalities).  However, OmniSat makes a key change to how it determines negative pairs, and denotes that any patches which neighbor the positive pair should not be considered as a negative pair.  This may seem odd from the perspective of natural imagery, but within RS imagery, there tends to not be vastly different semantic information in neighboring patches.  This follows Tobler's first law of geography: "Everything is related to everything else, but near things are more related than distant things" \cite{miller_toblers_2004}.  For the purposes of contrastive learning, it means that trying to use those neighboring patches as negative pairs will likely result in false negatives.  This contrastive strategy addresses that false negative issue.  OmniSat then performs independent masking, combines the latent representations into a multi-modal representation, and then reconstructs the original image modes from the shared multi-modal representation.  They choose MS timeseries reconstruction targets by considering the temporal attention maps.  Lower self-attention scores indicate cloudy images, which make for poor reconstruction targets.

\begin{figure}[ht]
    \centering
    \includegraphics[width=0.9\linewidth]{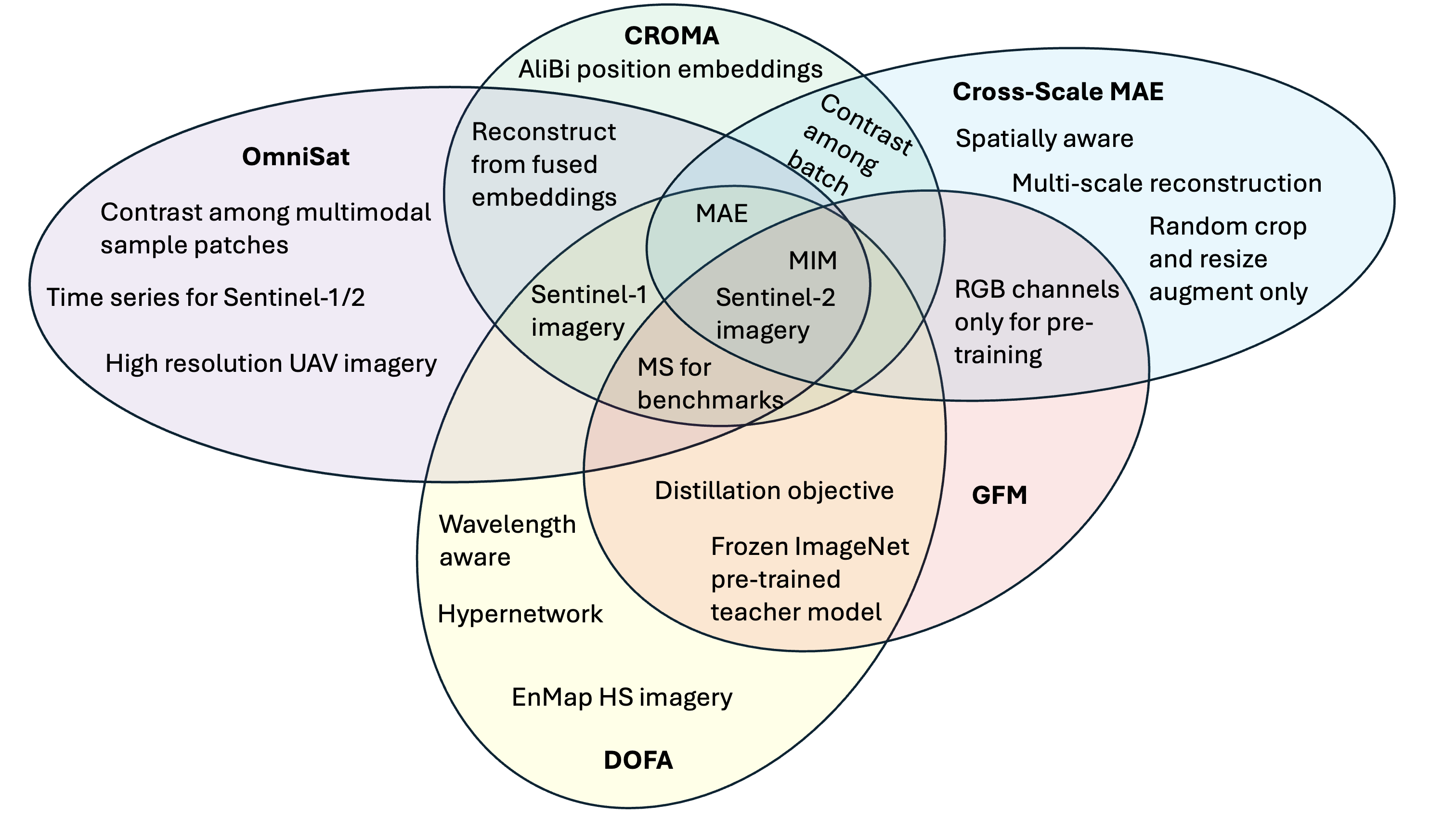}
    \caption{An illustration of shared and unique concepts among RS foundation models which utilize hybrid learning strategies that combine contrastive, MIM, and distillation approaches.}
    \Description{A sort of Venn diagram where a variety of RS models that utilize hybrid learning approaches are compared with one another to highlight similarities and differences.}
    \label{fig:hybrid_approaches_breakdown}
\end{figure}

Cross-Scale MAE \cite{tang_cross-scale_2024} also leverages both contrastive and MAE approaches.  Where OmniSat and CROMA use this combined approach to better train on multi-sensor data, Cross-Scale MAE does so in order to better understand spatial resolution.  Trained on RGB data from the FMoW \cite{christie_functional_2018} dataset, Cross-Scale MAE performs a random crop and resize augmentation on a sample, then compares that to the original image.  This comparison contains three components:
\begin{enumerate}
    \item contrastive loss between embeddings of the original image and augmented image
    \item mean squared error between the output of the attention block of the decoder for the original and augmented images
    \item reconstruction loss of the original image (from a patchified version of the original) and augmented image (from a patchified version of the augmented image)
\end{enumerate}

Contrastive loss between different spatial resolutions of the same image ensures that the model treats images with the same subject matter the same, regardless of size.  Mean Square Error (MSE) loss between the attention block outputs is designed to reinforce that relationship, and then standard reconstruction loss encourages learning intra-image semantics.  OmniSat performs very well on the RESISC45 \cite{cheng_remote_2017} task, which features images with varying spatial resolution, compared to other spatially aware models, such as ScaleMAE \cite{reed_scale-mae_2023}.  OmniSat explores adding the GSD embedding present in ScaleMAE and finds that it does not improve performance.  The author's theorize that it provides redundant information, and further point to the fact that incorporating GSD on SatMAE \cite{cong_satmae_2023}, which has no spatial awareness, improves performance.

Geospatial Foundation Model (GFM) \cite{mendieta_towards_2023} also leverages multiple approaches to train on a single sensor.  Where Cross-Scale MAE focuses on understanding spatial resolution, GFM focuses on treating RS imagery as an extension of natural imagery, performing continual pre-training with models trained on the ubiquitious ImageNet \cite{russakovsky_imagenet_2015} dataset.  GFM draws inspiration from the fact that that MIM can trivialize the task when training on data with low entropy.  In other words, if there is not much difference from patch to patch, then the network simply does not work as hard to reconstruct a masked patch. GFM calculates the average entropy of 3k randomly selected images from ImageNet and Sentinel-2 and find that Sentinel-2 images have an entropy of 3.9, compared to an entropy of 5.1 for ImageNet \cite{mendieta_towards_2023}.  GFM develops a two-pronged learning approach to address this issue.  GFM maintains a student encoder and teacher encoder, but the teacher encoder is an entirely frozen model initialized with weights from training on ImageNet.  For each image, the student is fed a masked version of the image, and the teacher is fed an unmasked version.  Then, there are two loss terms generated:
\begin{itemize}
    \item the cosine similarity between the teacher and the student embeddings
    \item reconstruction loss from trying to reconstruct the original image through the student embeddings
\end{itemize}

This framework is designed to leverage pre-exisiting work in CV via distillation to minimize the overhead needed to train RS foundation models.  Distilling against a pre-trained ImageNet model lets GFM quickly learn how to extract features from images, taking advantage of the higher entropy present in the ImageNet images.  The additional MIM task ensures that GFM learns RS domain-specific information.

Dynamic One-For-All (DOFA) \cite{xiong_neural_2024} builds on GFM's approach of distilling against a CV model pre-trained on ImageNet while also creating a framework that can take in data from any number of different sensors.  DOFA follows a hypernetwork approach \cite{chauhan_brief_2024}, where a hypernetwork takes in the wavelength details of a given band and then generates weights that are used to initialize a neural network.  Image bands are masked, then fed into that neural network to generate encodings.  The encodings are then fed through an encoder-decoder structure, with a mirrored hypernetwork at the end responsible for reconstructing the original image.  This framework is designed to learn bandwidth specific features via the hypernetwork and shared features via the shared encoder-decoder.  DOFA is trained on SAR imagery from Sentinel-1, multispectral imagery from Sentinel-2, hyperspectral imagery from enMAP, and high resolution RGB imagery from NAIP.  For the distillation objective, DOFA generates embeddings from the RGB channels (if present) and compares those embeddings to those of an ImageNet pre-trained teacher model on the same RGB bands.  In the case of SAR imagery, a random polarization is selected for the third channel to ensure the correct input shape for the CV teacher model.  Similar to GFM, the authors of DOFA cite a desire for reduced C02 emissions and efficiency during pre-training \cite{xiong_neural_2024} as a motivation for the distillation objective.  The resulting framework achieves SOTA performance on downstream tasks such as BigEarthNet \cite{sumbul_bigearthnet_2019}, So2Sat \cite{zhu_so2sat_2019}, and EuroSat \cite{helber_eurosat_2019} at the time of publication in 2024.

While this framework is highly promising in its potential applications to downstream tasks on imagery acquired via a sensor not in the training set, this ability is unfortunately still largely theoretical, as it is not thoroughly explored in the published results \cite{xiong_neural_2024}.

\section{Future Considerations}
Given the relative youth of foundation model development in RS, there is a high degree of fluidity in existing frameworks and overall goals.

\subsection{Learning Representations in the Era of Climate Change}
Given the rapidly shifting nature of the Earth's climate, learned representations should be robust enough to take these changes into account without substantial retraining.  Otherwise, the re-usability of these representations becomes less desirable and/or reliable over time, diminishing one of the main benefits of foundation models.  When considering the variety of modalities present in RS, there is the potential to create representations that can withstand the challenges of a changing climate, but a degree of care is required.  Relying too heavily on geographic locations and temporal data could very well result in representations that quickly become outdated as the world’s climate shifts.  For example, a learned representation of a specific location on a coastline could cease to be relevant due to rising sea levels.  Similarly, climate change has resulted in marked seasonal shifts within the past decade.  Therefore, learning representations of what a given location looks like at a given time of year may wane in relevance as well.

By comparison, representations based on modalities that directly observe the Earth, such as satellite imagery and text, should be more tolerant to climate change-induced drift.  Such representations are tied to understanding underlying observations directly within the data and are less reliant on climate context.  Given that many remotely sensed datasets date back decades, there is merit in creating foundation models that perform time-series forecasting over long time periods in order to effectively learn representations which take long-term trends into account.  This historical data also opens the door for assessing the capability of RS foundation models to quantify historically understood climate and biome changes.

The combination of historical data and the steady incoming stream of satellite data from currently orbiting satellite offers intriguing possibilities for continual pre-training, an approach popularized in NLP \cite{yildiz_investigating_2025}.  Continual pre-training focuses on efficiently adapting existing models to emerging knowledge in a field while maintaining previously learned knowledge.  Within the remote sensing domain, GFM \cite{mendieta_towards_2023} has begun to explore this approach.  However, GFM does so by continually training a CV model against RS data, instead of continually training an existing RS model against a stream of new RS data.  Developing a foundation model on historical data, and then performing continual pre-training against up-to-date RS data could prove useful for the development of climate change-tolerant RS foundation models.  This approach may prove particularly pertinent when working with spatiotemporally dependent data modes, where the same scene may be re-contextualized due to climate change.

\subsection{Under-Tapped Modes for Geospatial Representation Learning}
Many current RS foundation models build on CV foundation model frameworks.  Given the success of foundation models in the CV domain and the sheer amount of imagery that is available, be it from satellites or other aerial sources, this is a necessary first step. However, some models have achieved success by training heavily on non-image modes of data.  For example, MMCPP \cite{chen_mmcpp_2023} uses the CLIP contrastive learning objective and trains on population trajectory maps, points of interest, and aerial RGB images.  Such models make a compelling case to avoid purely focusing on satellite and/or aerial imagery when training RS foundation models.  

Environmental stations measure features such as groundwater, weather, or snow melt data.  They provide important climate and atmospheric information at a very high temporal resolution, often recording data on a daily or even hourly basis.  More human-centric data, such as social media posts, social connectedness indices \cite{bailey_social_2018}, and trajectory data, also provide invaluable information for understanding more societally-focused downstream tasks.  Both forms of information are not observable from space, but could provide useful modes due to their different perspective and high temporal resolution.  Data on rapidly changing phenomena also tends to not be observable from space.  Changes can occur too quickly for a satellite to capture due to orbit and revisit time.  GAIR's \cite{liu_gair_2025} usage of INR representations for satellite imagery also provides a useful starting point for mapping granular, ground-level data to satellite imagery.

\subsection{Performance versus Training Expenses}
Given the computational costs of training transformer models, the training expense of foundation models has rapidly become an obstacle across a variety of domains, with some models now including a section on carbon footprint \cite{cong_satmae_2023}.  As a substantial portion of downstream tasks for remote sensing pertain to combating the negative effects of climate change, there is a compelling argument for remote sensing foundation models to focus on curtailing training expenses.  Massive models which create a large carbon footprint ultimately may be counterproductive from a climate change standpoint.  However, the perpetual flow of remotely sensed data which dates back decades also encourages exploring the development of an incremental learning pipeline to efficiently fine-tune foundation models with up-to-date information.

Aside from a carbon footprint perspective, models with huge numbers of parameters have their downstream applications somewhat limited due to computational requirements.  Not all users have the hardware necessary to fine-tune foundation models with a high parameter count.  Some models, such as MOSAIKS \cite{rolf_generalizable_2021}, focus on lightweight frameworks in order to promote access to remote sensing foundation models.  MOSAIKS has competitive results on a number of downstream tasks such as forest cover prediction\cite{hansen_high-resolution_2013} at the time of publication in 2021, indicating that increasing parameters within RS foundation models may follow a law of diminishing returns. On the other hand, the encoder-decoder structure that is predominant in foundation models allows for repeated, diverse usage of representations on downstream tasks.  Arguably, leveraging significant resources to generate high quality embeddings as a one-time cost is the most effective path moving forward, as this approach results in SOTA performance on downstream tasks with relatively few resources necessary for future fine-tuning.

Other domains have specifically studied how the factors of data size, number of parameters, and compute affect overall model performance \cite{kaplan_scaling_2020}.  NLP has established a principle that performance increases with an increase in any of those factors, providing that neither of the other two serve as a bottleneck \cite{kaplan_scaling_2020}.  However, such studies in CV for MIM have demonstrated that large-scale data is a requirement for scaling up compute and model parameters \cite{xie_data_2023}.  Moreover, MIM cannot benefit from increasing the data size if the model is not overfitting \cite{xie_data_2023}.  To the best of our knowledge, such studies do not exist for the remote sensing field.  Authors in CV cite the semantic richness in text tokens compared to the redundancy of visual tokens as a significant motivator for their study \cite{xie_data_2023}.  It stands to reason that the significant differences in natural imagery and remotely sensed imagery would also motivate such a study in the remote sensing field.  

It is possible that SSL distillation networks may provide a happy medium between the desire for high quality embeddings and lower computational requirements.  While models such as RS-BYOL \cite{jain_self-supervised_2022}, DINO-MM \cite{assran_self-supervised_2023}, and SkySense \cite{guo_skysense_2024} have demonstrated that such an approach is feasible with remotely sensed imagery, distillation has yet to be explored as a SSL approach in remote sensing to the degree that masked image modelling or contrastive learning with negative samples have been.  The recent and well-publicized success of DeepSeek \cite{wu_deepseek-vl2_2024} in the NLP field also highlights the viability of distillation for knowledge compression and indicates that it should be further examined within the remote sensing domain.  GFM \cite{mendieta_towards_2023} begins to inspect the efficacy of distilling knowledge from the CV domain into remote sensing, but there remains substantial room for exploration.

A Mixture of Experts (MoE) approach may also provide an intriguing compromise between embedding quality and computational requirements.  At a high level, MoE partitions an input space into non-overlapping regions, then designates an 'expert' model per region \cite{gamal_moe_2025}.  This particular approach allows for flexible incorporation of models for a variety of modalities and tasks, yet remains computationally efficient since only a subset of experts are activated at training and inference.  The Switch Transformer \cite{fedus_switch_2022}, an NLP model for multilingual learning, followed the MoE paradigm to develop a foundation model with over a trillion trainable parameters.  In addition to the enormous parameter count, the Switch Transformer also trains significantly faster than comparable models on the same hardware \cite{fedus_switch_2022}.  The MoE framework's capabilities in efficiently leveraging mode and task specific models make it an exciting avenue for exploration in RS foundation models.  To the best of our knowledge, a MoE-based remote sensing foundation model does not yet exist.

\subsection{Directions for Framework Development}
Many of the foundation models covered in this manuscript leverage a variety of different objectives that force the encoder to learn accurate representations of the data and test on downstream generalizability.  Many of these models use a single pretext task training objective in the hopes of generating a general representation.  In other domains, such as NLP, multiple pretext tasks are optimized simultaneously as a key part of learning generalizable embeddings \cite{koroteev_bert_2021}.  RS foundation models could profit from a similar strategy.

MMEarth \cite{nedungadi_mmearth_2024} utilizes multi-task training, as the model uses a variety of downstream decoders for different modalities that each backpropagate loss to the same encoder.  FG-MAE \cite{wang_feature_2023} takes this a step further by tasking decoders to not just reconstruct a different modality, but to reconstruct specific RS feature maps pertaining to the masked training image.  While we have discussed some approaches designed to leverage contrastive loss between modes that also include MAE within a mode, this remains an under-explored avenue.  CROMA \cite{fuller_croma_2023} begins to explore this possibility in the RS domain by contrasting two independently masked images, but this is explored mainly for performance reasons, and the MAE objective is on multi-sensed image embeddings as opposed to MAE for each sensor.  This approach may scale poorly with the addition of more sensors due to input size.  Similarly, the nature of a MAE task shared with multiple modalities seems to align these sensors more closely, rendering it somewhat redundant to the contrastive task already being executed.  DeCUR \cite{wang_decoupling_2024} provides an interesting alternative to such approaches.  It leverages the same redundancy reduction task across modes and within modes, but the choice to set aside roughly 20\% of the latent space dimensions for learning modality specific features seems to be a framework choice that other foundation models could benefit from.  Similarly, the regularization term in VICReg \cite{bardes_vicreg_2022} that enforces a degree of standard deviation within each dimension is worth exploring as an avenue to prevent embedding collapse.

Most multi-modal models in the RS domain only consider two to three sensors, but the sheer number of modes available means that it would be beneficial to have a model that is capable of scaling effectively to learn representations that capture a dynamic number of modes.  MMEarth’s innovation of having a decoder per mode is promising in that regard, but it runs into the problem of only learning embeddings for one sensor.  While contrastive learning can theoretically be applied across a variety of modes, the cost of training a separate encoder per sensor could quickly result in resource challenges.  For example, OmniSat \cite{astruc_omnisat_2024} leverages MAE per mode and contrastive loss across three modes, but the contrastive framework would scale poorly with the addition of more modalities. Ultimately, a multi-modal framework that is capable of learning representations for multiple sensors without explicitly requiring that every mode has its own encoder would be ideal, but careful consideration is required to develop a framework that is both flexible and robust enough.  One potential option would be following an ImageBind \cite{girdhar_imagebind_2023} approach, where one modality is the 'binding' modality that all other modalities are contrasted to.  Treating spatiotemporal location as a binding modality for any remotely sensed data could be an interesting avenue to pursue.  Such a framework would allow incorporation of pre-trained, sensor-specific encoders.  The number of comparisons made would scale linearly with the number of additional modalities, as opposed to exponentially.

DOFA's \cite{xiong_neural_2024} hypernetwork-inspired framework is promising in that regard due to its ability to handle a variety of different sensor inputs in pre-training, but it fails to take into account whether a wavelength is being actively sensed or passively sensed, which affects the nature of the imagery. Similarly, the various polarization of SAR acquisitions, even by the same sensor, can create further complications that go beyond wavelength: The C-band Sentinel-1 SAR images land areas in VH and VV mode, whereas it switches to HH and HV polarizations when over oceans (this is driven by the scientific and operational users' need to map sea ice, which is better detectable in these modes).  DOFA also does not demonstrably test what we consider to be this framework's most interesting potential capability: the capacity to operate effectively on downstream tasks that rely on a sensor not found in the training dataset.

SeCo \cite{manas_seasonal_2021} and GASSL \cite{ayush_geography-aware_2021} demonstrated the value of utilizing temporal data in training, but cautioned that the learned representations were agnostic to seasonal changes.  The seasonal contrastive objective could be helpful as an additional task, when taken into account with other pre-training tasks that take seasonal change into account.  Ultimately, multi-task training on complementary tasks, such as this particular example, is somewhat costly, but it results in representations that learn the unique features inherent to each task while avoiding the problem of overfitting.  Aside from the SeCo objective,  SatMAE’s \cite{cong_satmae_2023} breakthrough in adding temporal information to the positional encoding of images could serve as a useful way to introduce temporal information into existing methods.  However, caution should be taken to ensure that such encoders are robust enough to not be overly reliant on temporal information.  ScaleMAE’s \cite{reed_scale-mae_2023} novel approach of normalizing positional embeddings to account for image scale also seems to be a methodology that could be easily applied to other models within the field, presuming those models utilize a ViT framework.  Similarly, Cross-Scale MAE's \cite{tang_cross-scale_2024} focus on random crop and resize augmentations as a way to learn spatial resolution could be applied to other contrastive and distillation based frameworks.

\section{Conclusion}
In this paper we presented on a variety of SSL approaches currently utilized in RS foundation models.  We traced these approaches back to their origins in the CV field to develop a better understanding of the current swathe of RS foundation models and potential paths for future development. 

We examined:
\begin{enumerate}
    \item Varying benefits and drawbacks of RS embeddings learned with different frameworks and SSL tasks.
    \item Prospective options for lowering compute costs for training RS foundation models.
    \item Inherent opportunities available for leveraging multi-sensory Earth observations to train RS foundation models on publicly available unlabeled data.
\end{enumerate}

\begin{acks}
To the GeoHAI lab, for all of their ideas, support, and kindness.
\end{acks}

\bibliographystyle{ACM-Reference-Format}
\bibliography{references}

\end{document}